\documentclass[letterpaper, 10 pt, conference, final]{ieeeconf}  

\IEEEoverridecommandlockouts                              

\usepackage{mathptmx} 
\usepackage{amsmath} 
\usepackage{amssymb}  
\usepackage{siunitx}
\usepackage{url}
\usepackage[shortcuts]{extdash}
\usepackage{float}
\usepackage{booktabs}
\usepackage{tikz}
\usepackage[utf8]{inputenc}
\usepackage{pgfplots}
\DeclareUnicodeCharacter{2212}{−}
\usepgfplotslibrary{groupplots,dateplot}
\usetikzlibrary{patterns,shapes.arrows}
\pgfplotsset{compat=newest}
\usepackage{subfig}
\usepackage{circledsteps}
\usepackage{graphicx}
\usepackage{fancyhdr}
\title{\LARGE \bf AIMY: An Open-source Table Tennis Ball Launcher for\\Versatile and High-fidelity Trajectory Generation}

\author{Alexander Dittrich, Jan Schneider, Simon Guist, Nico Gürtler, Heiko Ott, Thomas Steinbrenner, \\Bernhard Schölkopf and Dieter Büchler
\thanks{This work was supported by the Max Planck Institute for Intelligent Systems.}
\thanks{All authors are affiliated with the MPI for Intelligent Systems, Max-Planck-Ring 4, 72076 Tübingen, Germany.}%
}

\begin{document}

\maketitle
\thispagestyle{fancy}
\begin{abstract}
To approach the level of advanced human players in table tennis with robots, generating varied ball trajectories in a reproducible and controlled manner is essential. 
Current ball launchers used in robot table tennis either do not provide an interface for automatic control or are limited in their capabilities to adapt speed, direction, and spin of the ball. 
For these reasons, we present AIMY, a three-wheeled open-hardware and open-source table tennis ball launcher, which can generate ball speeds and spins of up to \SI{15.4}{\meter\per\second} and \SI{192.0}{\per\second}, respectively, which are comparable to advanced human players. 
The wheel speeds, launch orientation and time can be fully controlled via an open Ethernet or Wi-Fi interface. 
We provide a detailed overview of the core design features, and open-source the software to encourage distribution and duplication within and beyond the robot table tennis research community.
We also extensively evaluate the ball launcher's accuracy for different system settings and learn to launch a ball to desired locations.
With this ball launcher, we enable long-duration training of robot table tennis approaches where the complexity of the ball trajectory can be automatically adjusted, enabling large-scale real-world online reinforcement learning for table tennis robots.
\end{abstract}
\section{Introduction}

Table tennis is a fast-changing and uncertain challenge for robots, where current systems do not reach professional human performance. In modern competitive table tennis, effective command of ball speed and spin is crucial both for offensive and defensive play. While current state-of-the-art table tennis robots can cope with increasing reliability with medium to high-paced shots, ball trajectories with considerable spin remain a significant obstacle on the way to reach the “AlphaGo moment” of robot table tennis.

Recent successes in robot table tennis~\cite{tebbe2021, luo2021, buechler2022, abueyruwan2022} show that Reinforcement Learning~(RL) has the potential to significantly increase the performance of table tennis robots. However, training with large-scale real-world interactions of robot and ball -- as required with RL -- is hard to carry out practically and hence these approaches rely on simulations in some form. Simulating contacts between racket and ball and in the bounce with the table require an increasingly complex rebound model, which is more difficult to identify with increasing speed and spin of the ball. This limitation restricts progress in achieving performance comparable to advanced human players.

Consequently, rich real-world data created during training under competitive conditions holds great potential for acquiring even more sophisticated table tennis skills. For meaningful use of this data in sample-intensive learning, replications of a wide range of challenging but reproducible game situations are of great importance. Therefore, a reliable method is needed to generate versatile, high-fidelity ball trajectories.

\begin{figure}[t]
  \centering
  \includegraphics[width=0.48\textwidth]{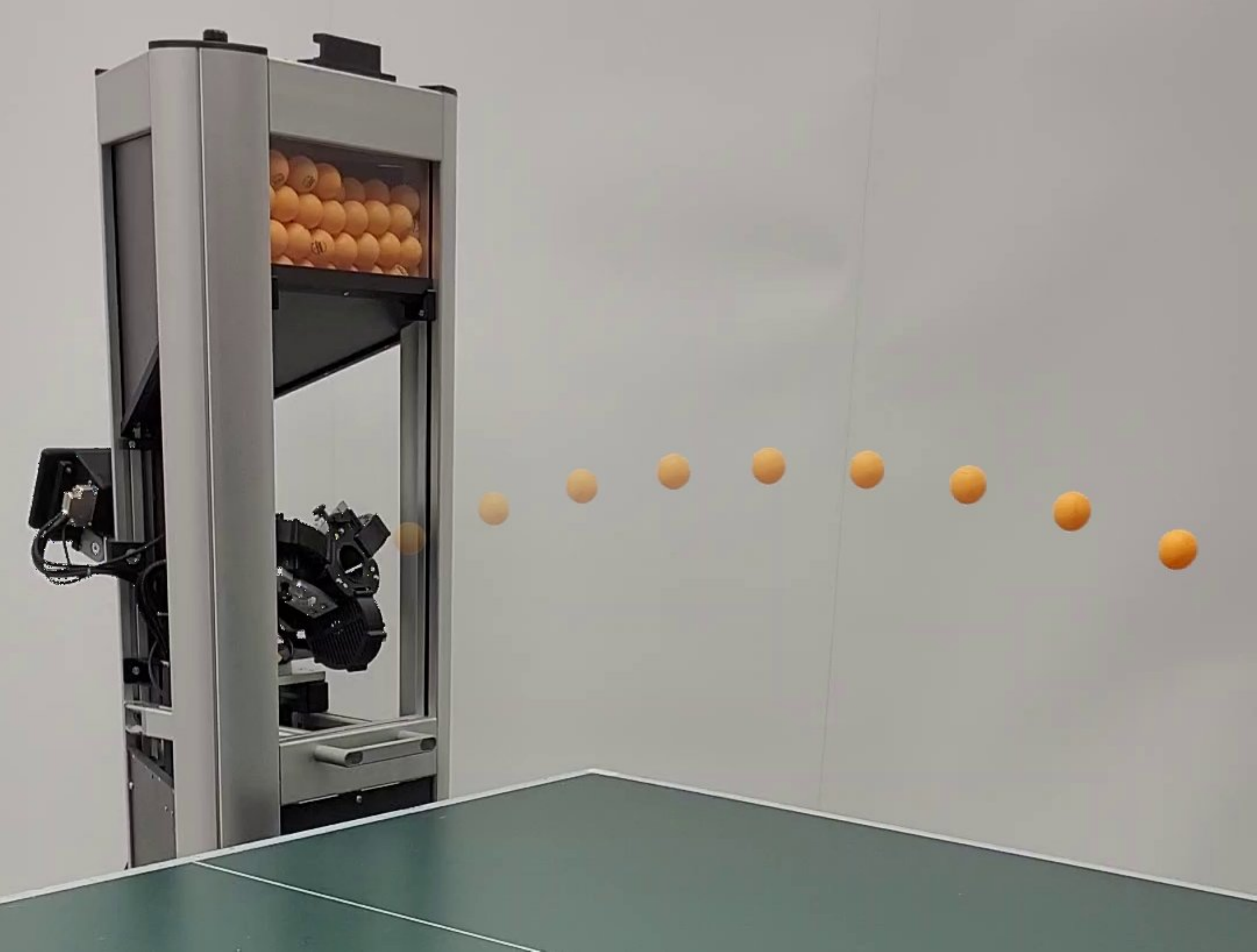}
  \caption{AIMY, an open-source table tennis ball launcher capable of generating speeds and spins comparable to advanced human players.
  The five independent degrees of freedom allow launching balls with arbitrary target positions, orientations, and spins.}
  \label{fig:launcher}
\end{figure}

For this purpose, we developed the high-precision table tennis ball launcher that we name AIMY and depict in Figure~\ref{fig:launcher}. In experiments, we show that AIMY reliably generates arbitrary trajectories with translational speeds up to \SI{15.4}{\metre\per\second} and spins of approximately \SI{192.0}{\per\second} with high accuracy. For comparison, Lee et al. \cite{lee2019} measured average ball speeds of \SI{16.8}{\metre\per\second} and ball spins with \SI{113.9}{\per\second} among players on national table tennis team level, which used maximum strength to hit a defined target.

To make AIMY as adjustable and easy to use as possible, the user can adjust the wheel velocities and the orientation of the ball launch unit via a remote low-latency Ethernet interface. We provide a Python API to modify these parameters and enable controlled and precisely timed launching.

The contribution of this paper is four-fold: We (i) open-source the hardware and the software interface\footnote{AIMY project web page: https://webdav.tuebingen.mpg.de/aimy/},
(ii) quantify the accuracy of our new ball launcher under varying hardware considerations,  
(iii) learn to return a ball to a desired landing position on the table with a supervised learning approach and
(iv) provide a rich and big data set of 3761 ball trajectories of various ball speeds and spins for benchmarking model learning approaches.
In this manner, we hope to facilitate research on more complex scenarios in robot table tennis.

In the following, we provide an overview of the core design features and components, along with the corresponding low-level control software, in Section~\ref{sec:design}. Subsequently, Section~\ref{sec:accuracy} reports the evaluations on repetition accuracy for different system parameters. Furthermore, AIMY is validated for different target distances compared to commercially available products. Section~\ref{sec:target} describes a supervised learning approach to target shooting, as well as the data set of various ball trajectories for model learning.
\section{Related Work}
The majority of past works in robot table tennis focussed on data-efficient approaches to return table tennis balls of lower speeds and less spin compared to professional play, and therefore did not require an automated ball launcher. 
For example, multiple studies use real-world data on a small scale such as in Matsushima et al.~\cite{Matsushima2005}, Huang et al.~\cite{huang2016jointly}, Ma et al.~\cite{malearning}, Koç et al.~\cite{koc2018}, Gomez-Gonzalez et al.~\cite{gomez2020adaptation} and Mülling et al.~\cite{muelling2013} that either use some form of iterative learning control or imitation learning approaches. All these approaches required only small sample sizes. 
More recent approaches that use RL, avoid long-term training with real balls by using prerecorded data and train in a hybrid sim and real setting~(Büchler et al.~\cite{buechler2022}), use data efficient approaches and reduce the task to its minimum complexity~(Tebbe et al.~\cite{tebbe2021}), or leverage sophisticated simulation environments~(Yang et al.~\cite{yang2021} and Abeyruwan et al.~\cite{abeyruwan2022}).

A large number of commercially available table tennis launchers exist and are tailored for training with humans ranging from beginners to advanced players but not for automated long-term training. The TTmatic 303A~\cite{ttmatic303}, for example, accelerates the ball with a single wheel and a manually exchangeable slide that alters the altitude of the launching angle. In an azimuthal motion, the launching frequency as well as the motor speed are adapted via a physically wired control panel. Usage of this ball launcher for long-term training is not feasible since launching parameters and launching frequency have to be changed manually.
Additionally, it cannot adapt the spin of the ball. Other models like the Donic Newgy Robo-Pong~2050~\cite{robopong3050xl} feature the manual rotation of the single wheel launch unit to enable restricted spin generation. More sophisticated products like the PowerPong Table Tennis Robot series~\cite{powerpongomega} and the Butterfly~AMICUS series~\cite{amicusprime} rely on a three-wheel design concept to generate noteworthy spin. However, neither of these products provide an API that allows researchers to control launching parameters and launching. 

The number of reports on custom-designed ball launchers in the scientific literature is fairly limited. For example, \cite{nemire1991, ponnusamy2006} facilitate automated devices for table tennis ball launching, but neither of them satisfies the need for remote controlled, high-fidelity ball trajectory generation.

\section{Design Overview}
\label{sec:design}
Multiple objectives guided the development of the ball launcher. 
This section summarizes them, followed by the presentation of the core mechanical design features and the software framework.

\subsection{Design Objectives}
The main intention for creating the ball launcher is to generate versatile ball trajectories in a controlled and repeatable fashion to facilitate research in robot table tennis.
The following list contains detailed and specific derivations of this main goal that steered the development of AIMY:
\begin{itemize}
    \item \textbf{Arbitrary Trajectory Generation (R1):} Sufficient coverage of target area on an International Table Tennis Federation (ITTF) compliant table tennis table; capability to generate ball speed (approx. \SI{16}{\metre\per\second}) and spin (approx. \SI{114}{\per\second}) comparable to professional human level.
    \item \textbf{High-fidelity Launching (R2):} Sufficiently repetitive, low-deviation target point hitting; accuracy should be equal or better than commercially available products; low number of trajectories with large deviation (outliers) for a fixed parameter setting.
    \item \textbf{Suitability for Long-duration Usage (R3):} Characteristics of the ball trajectory should remain the same for the same parameters also under long-duration measurement conditions; reliable supply of the launch unit without clogging of ball supply; reliable launching without interrupts by the launch unit.
    \item \textbf{Remote Controllability (R4):} Remote online parameter control in soft real-time (approx. \SI{500}{\milli\second}); remote on demand ball launching in soft real-time (approx. \SI{500}{\milli\second}); launching also in irregular intervals, e.g., due to computation time of policy optimization after epochs.
    \item \textbf{Advanced Trajectory Specification (R5):} Automated launch parameter selection for specification of landing position, angle of ball impact, ball speed, and ball spin. 
    \item \textbf{Costs (R6):} Material and component costs in total should not significantly exceed purchase prices of commercial products and should provide a good compromise between the necessary accuracy, longevity, and price.
    \item \textbf{General Handling Aspects (R7):} Manufacturability with regular workshop equipment; assemblability; maintainability; extensibility; user-friendly installation and mechanical parameter adjustment.
\end{itemize}

\subsection{Hardware Overview}
\label{sub:hardware}
\begin{figure*}[th]
    \centering
    \subfloat[]{
        \includegraphics[width=10.0cm]{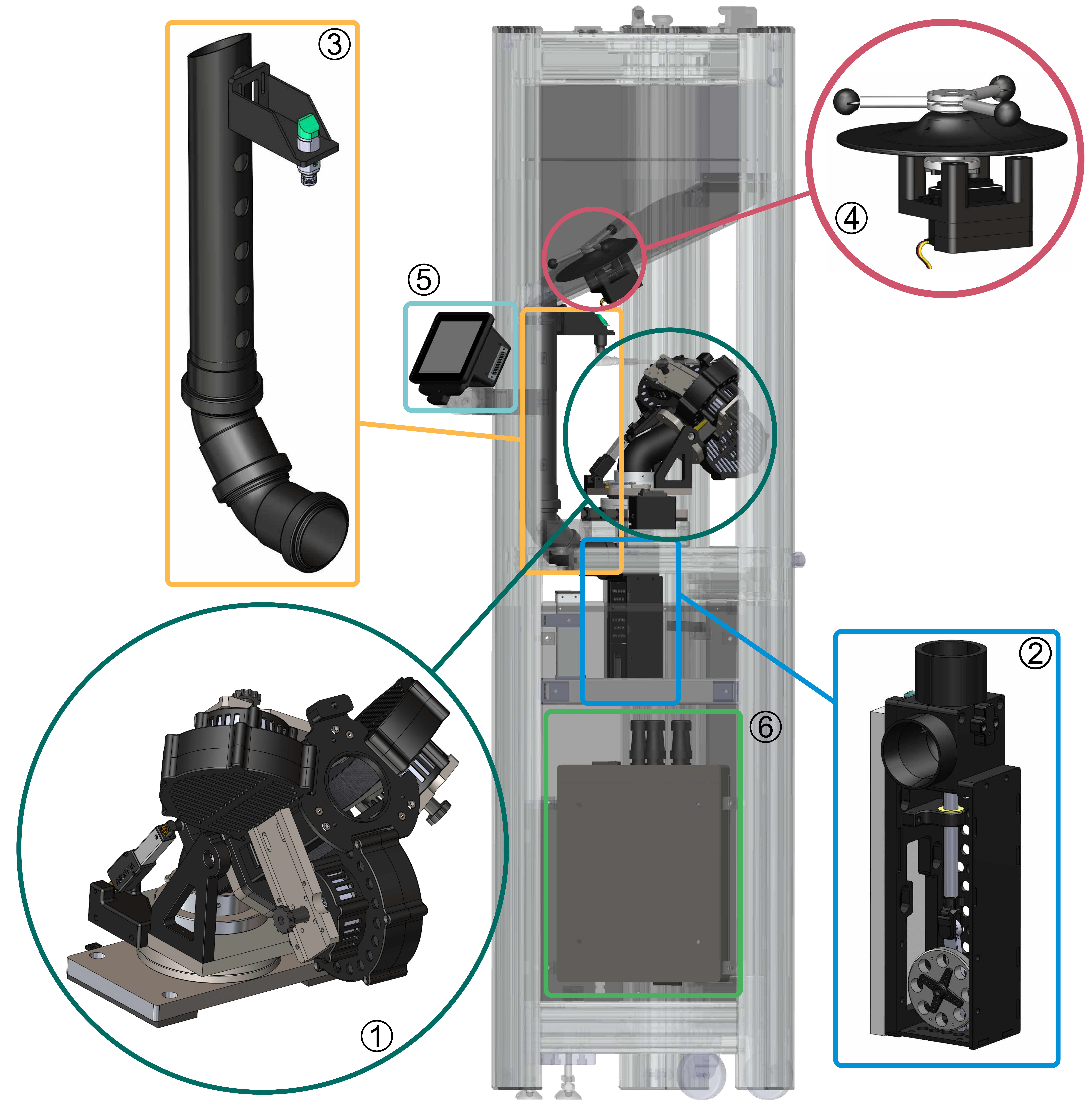}
        \label{fig:components}
    }
    \hfill
    \subfloat[]{
        \includegraphics[width=7.0cm]{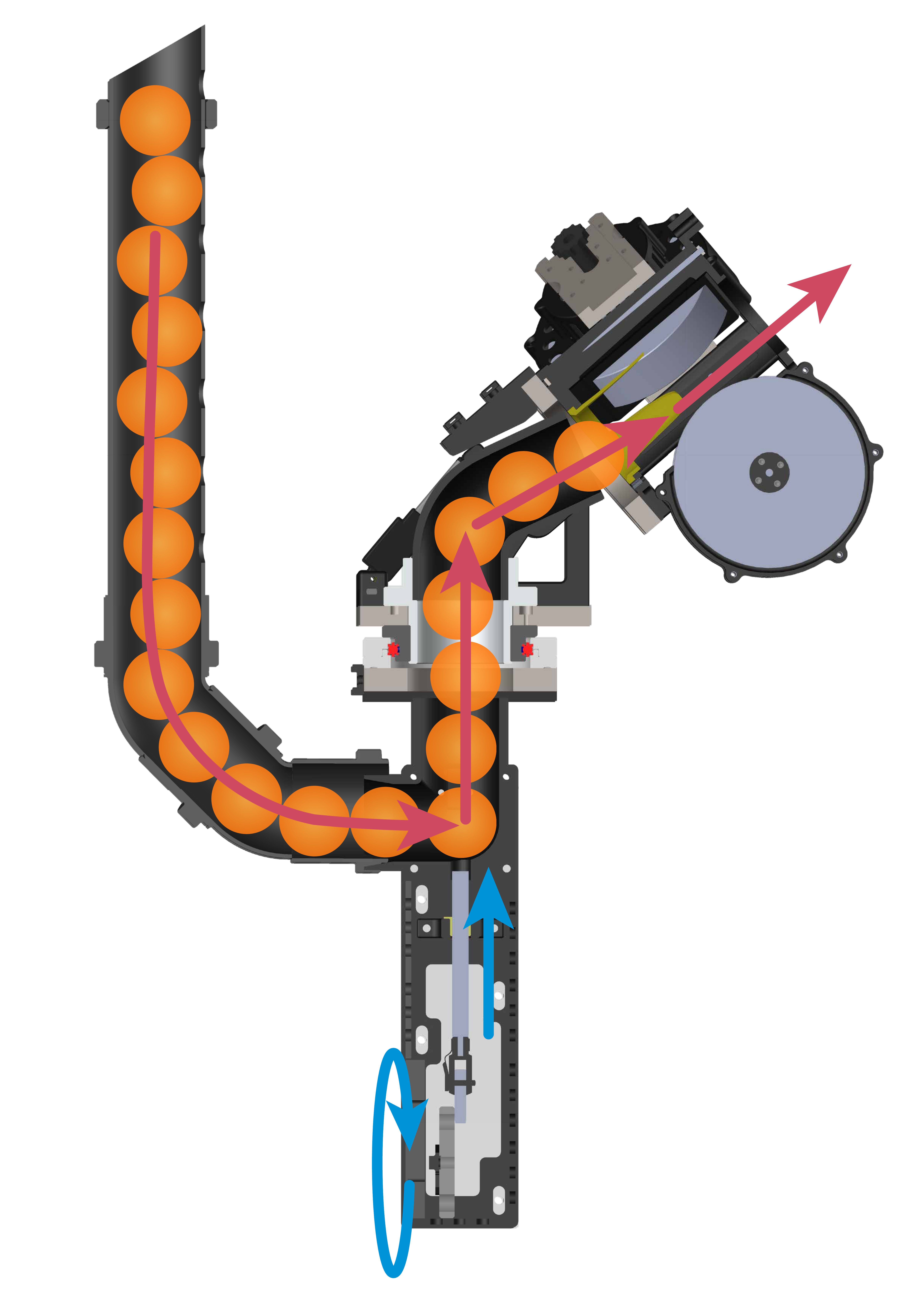}
        \label{fig:feeding_system}
    }
    \caption{\textbf{(a)} Overview of AIMY with core functional units. \Circled{1} Launch unit in three-wheel configuration. The launch unit is rotatable due to a ball bearing in combination with a servo motor and adjustable in altitude angle via a linear motor. \Circled{2} Ball feeding unit with crank mechanism for supplying the launch unit with balls from the ball supply channel. \Circled{3} Ball supply channel connecting ball supply unit with ball reservoir. The ball supply channel is equipped with an optical sensor for detecting the state of filling. \Circled{4} Reservoir stirrer generates movement in the ball reservoir for resolving ball clogging. \Circled{5} The control unit enables hardware control, hosts the ball launcher server for remote controllability, and provides a graphical user interface via a 7-inch touchscreen. \Circled{6} Electronics box with a 5~V and 24~V power supply and ESCs for motor control.
    \textbf{(b)} Sectional view of the ball feeding flow from reservoir to launch unit. Ball supply mechanics from ball supply channel through ball feeding unit towards launch unit and launch, shown in a cross-section view of the respective components. Red arrows indicate ball flow, blue arrows indicate flow of forces via the crank mechanism.}
\end{figure*}

With these design objectives in mind, AIMY is divided into distinct functional core units, which are depicted in Figure~\ref{fig:components}. The most important of these core units are the launch unit, the ball feeding system consisting of the ball feeding unit and the ball supply channel, and the reservoir stirrer. 

\subsubsection{Ball Feeding System}
The ball feeding system ensures that the launch unit is supplied with table tennis balls from the ball reservoir. The ball feeding unit uses a crank mechanism to transform the rotary motion of the servomotor (Hitec HSB-9381TH) into a linear motion, pushing the balls in queue into the engagement point of the throwing wheels of the launch unit. Figure~\ref{fig:feeding_system} illustrates this process. Furthermore, an optical sensor (Pepperl+Fuchs GLV18-8-400-S) mounted at the ball supply channel detects the state of filling and triggers the reservoir stirrer if required.

\subsubsection{Launch Unit}
The launch unit receives a ball from the ball feeding system and accelerates it via three throwing wheels that are arranged at \ang{120} to each other. Inspired by commercial products like the Butterfly~AMICUS~Prime or the PowerPong~Omega, this concept can generate significant amounts of spin around the launching tube (yaw) and the horizontal ball axis (pitch). Roll spin can be produced to a limited extent via suitable selection of orientation and wheel speed parameters. For actuating the wheels, the high-performance brushless motor Antigravity~MN5008~KV170 of the manufacturer T\=/MOTOR has been selected. Although developed for drone propulsion, this series of motors proved to be reliable in other robotics applications, such as in~\cite{wuthrich2020} and~\cite{grimminger2020}. 
Hard foam wheels by the manufacturer Butterfly -- also used in the AMICUS table tennis launcher series -- transfer the motor torque to the ball.
The compliant body and the roughened surface reduces slip between the ball and wheel. The T\=/MOTOR~AIR~40A electronic speed controller supplies the motors of the launch unit. 
A regular 5~V servomotor (Hitec D954SW 32-bit) adapts the azimuthal orientation, allowing the launch unit to rotate by \ang{17} in both positive and negative direction. The altitude orientation of the launch unit can be modified via a linear actuator (Actuoinx L12-30-210-6-R) enabling angles from \ang{6.5} to \ang{37.0}. The selected linear actuator provides higher holding torque than comparable rotary   servo motors and is therefore better suited for the intended use.

\subsubsection{Reservoir Stirrer}
One characteristic of table tennis balls is a rough surface, which players leverage to their advantage by creating spin with rubberized table tennis rackets. 
However, this surface also promotes clogging of the balls in the ball reservoir. 
This issue results in the launch unit no longer being supplied with balls, which potentially disrupts the operation of the ball launcher and ends long-term tests. 
Therefore, in consideration of objective R3, a reservoir stirrer is added. The stirrer moves the balls in the reservoir to resolve potential clogging.
The motion is either started automatically after each launch or optionally by the signal of the sensor of the ball supply channel. A servo motor (Pololu SpringRC SM-S4303R) puts the stirrer into motion and can be equipped with different stirring attachments.

\subsubsection{Control Unit}
A network-compatible terminal device adjusts the hardware remotely. 
For this purpose, a Raspberry~Pi~4~Model~B with Raspberry Pi OS is used, since it offers a wide range of network interfaces but also enough GPIOs to directly address sensors and actuators. 
Xinabox OC05 servo drivers directly connected to the Raspberry~Pi via the GPIOs control the servomotors for the orientation of the launch unit, the reservoir stirrer and the ball feeding unit. 

\subsection{Software Framework}

The software is structured in a client-server scheme, where the control unit with direct access to the hardware hosts a server. Any device connected to the network can start a client for remote controlling the hardware. 
The server provides on-request functions for changing the current system state, such as orientation of the launch unit or wheel speeds, and can initiate launching at a desired time. 
While the server waits for control requests, it observes the state of the ball queue within the supply channel and activates the reservoir stirrer if necessary. The client software provides functions to submit requests to the server to initiate changes of the ball launcher's parameters and to initiate launching. For the client we provide a Python and C++ API, to enable embedding, e.g., in an RL environment. Server and client can be connected via regular Ethernet (wired) or Wi-Fi (wireless) networks. For communication, the open-source and widely supported messaging library ZeroMQ is used in a TCP configuration. Additionally, we provide a simple to use Tkinter graphical user interface for manual operation and testing of AIMY on a Raspberry Pi 7-inch touchscreen.
\section{Design and Accuracy Evaluation}
\begin{figure}[t]
    \centering
    \vspace{0.4cm}
\begin{tikzpicture}

\definecolor{black34}{RGB}{34,34,34}
\definecolor{dodgerblue1147215}{RGB}{1,147,215}
\definecolor{gray}{RGB}{128,128,128}
\definecolor{indianred20783105}{RGB}{207,83,105}
\definecolor{lightgray204}{RGB}{204,204,204}
\definecolor{teal0108102}{RGB}{0,108,102}
\definecolor{whitesmoke}{RGB}{245,245,245}

\begin{groupplot}[group style={group size=1 by 3, vertical sep=1.4cm}]
\nextgroupplot[
axis background/.style={fill=whitesmoke},
axis line style={gray},
height=3.5cm,
legend cell align={left},
legend style={fill opacity=0.8, draw opacity=1, text opacity=1, draw=lightgray204, fill=whitesmoke},
tick align=outside,
tick pos=left,
width=8.0cm,
x grid style={white},
xlabel=\textcolor{black34}{\(\displaystyle t_{\textrm{ramp}}\) [s]},
xmajorgrids,
xmin=-0.5, xmax=15,
xtick style={color=black34},
y grid style={white},
ylabel=\textcolor{black34}{\(\displaystyle \sigma\) [mm]},
ymajorgrids,
ymin=11.8378456058421, ymax=45.1120730937024,
ytick style={color=black34}
]
\addplot [line width=0.48pt, indianred20783105, mark=*, mark size=2.4, mark options={solid}]
table {%
0.01 40.1689226222973
0.05 40.0360561215795
0.1 43.5996082078906
0.5 21.1268140636482
1 24.2869301473277
2 24.3445778226545
3 23.4409623199056
8 23.7327250540187
};
\addlegendentry{$\sigma_{x,\textrm{ramp-up}}$}
\addplot [line width=0.48pt, dodgerblue1147215, mark=*, mark size=2.4, mark options={solid}]
table {%
0.01 16.5665564742826
0.05 14.3320032042472
0.1 17.4283277285046
0.5 15.0007166933308
1 17.1740209751143
2 13.4113465678336
3 13.3503104916539
8 17.2763155274512
};
\addlegendentry{$\sigma_{y,\textrm{ramp-up}}$}
\addplot [line width=0.48pt, teal0108102, mark=*, mark size=2.4, mark options={solid}]
table {%
0.01 28.36773954829
0.05 27.1840296629134
0.1 30.5139679681976
0.5 18.0637653784895
1 20.730475561221
2 18.8779621952441
3 18.3956364057798
8 20.504520290735
};
\addlegendentry{$\sigma_{\textrm{avg}}$}
\addplot [line width=0.48pt, indianred20783105, dashed, forget plot]
table {%
8 23.7327250540187
35 23.1068482921517
};
\addplot [line width=0.48pt, dodgerblue1147215, dashed, forget plot]
table {%
8 17.2763155274512
35 15.0455328140189
};
\addplot [line width=0.48pt, teal0108102, dashed, forget plot]
table {%
8 20.504520290735
35 19.0761905530853
};

\nextgroupplot[
axis background/.style={fill=whitesmoke},
axis line style={gray},
height=3.5cm,
legend cell align={left},
legend style={fill opacity=0.8, draw opacity=1, text opacity=1, draw=lightgray204, fill=whitesmoke},
tick align=outside,
tick pos=left,
width=8.0cm,
x grid style={white},
xlabel=\textcolor{black34}{\(\displaystyle p_{\textrm{stroke}}\) [-]},
xmajorgrids,
xmin=-1, xmax=32,
xtick style={color=black34},
y grid style={white},
ylabel=\textcolor{black34}{\(\displaystyle \sigma\) [mm]},
ymajorgrids,
ymin=11.8378456058421, ymax=45.1120730937024,
ytick style={color=black34}
]
\addplot [line width=0.48pt, indianred20783105, mark=*, mark size=2.4, mark options={solid}]
table {%
0.05 19.6968471252269
0.1 15.6307345068363
0.5 24.0627585301564
1 21.2650801017897
3 23.661425842816
5 22.6517158350676
10 19.8677397324046
30 27.9296133873545
};
\addlegendentry{$\sigma_{x,\textrm{stroke}}$}
\addplot [line width=0.48pt, dodgerblue1147215, mark=*, mark size=2.4, mark options={solid}]
table {%
0.05 14.2923178990428
0.1 19.2354820838708
0.5 23.906328791735
1 22.2961313151983
3 19.6001743191984
5 22.3218273192108
10 21.5241350461301
30 14.9931277190834
};
\addlegendentry{$\sigma_{y,\textrm{stroke}}$}
\addplot [line width=0.48pt, teal0108102, mark=*, mark size=2.4, mark options={solid}]
table {%
0.05 16.9945825121348
0.1 17.4331082953535
0.5 23.9845436609457
1 21.780605708494
3 21.6308000810072
5 22.4867715771392
10 20.6959373892673
30 21.461370553219
};
\addlegendentry{$\sigma_{\textrm{avg}}$}

\nextgroupplot[
axis background/.style={fill=whitesmoke},
axis line style={gray},
height=3.5cm,
legend cell align={left},
legend style={fill opacity=0.8, draw opacity=1, text opacity=1, draw=lightgray204, fill=whitesmoke},
tick align=outside,
tick pos=left,
width=8.0cm,
x grid style={white},
xlabel=\textcolor{black34}{\(\displaystyle d_{\textrm{pinch}}\) [mm]},
xmajorgrids,
xmin=35.135, xmax=38.765,
xtick style={color=black34},
y grid style={white},
ylabel=\textcolor{black34}{\(\displaystyle \sigma\) [mm]},
ymajorgrids,
ymin=11.8378456058421, ymax=45.1120730937024,
ytick style={color=black34}
]
\addplot [line width=0.48pt, indianred20783105, mark=*, mark size=2.4, mark options={solid}]
table {%
35.3 16.0566045415551
35.8 20.2740263366794
36.4 19.0314113856567
37 19.6486109356218
37.4 18.9861585942862
38.6 23.6136857453632
};
\addlegendentry{$\sigma_{x,\textrm{pinch}}$}
\addplot [line width=0.48pt, dodgerblue1147215, mark=*, mark size=2.4, mark options={solid}]
table {%
35.3 27.4695808719725
35.8 26.4265727361768
36.4 21.4700468292318
37 24.8010675187072
37.4 22.2645586213726
38.6 24.2818106138511
};
\addlegendentry{$\sigma_{y,\textrm{pinch}}$}
\addplot [line width=0.48pt, teal0108102, mark=*, mark size=2.4, mark options={solid}]
table {%
35.3 21.7630927067638
35.8 23.3502995364281
36.4 20.2507291074442
37 22.2248392271645
37.4 20.6253586078294
38.6 23.9477481796071
};
\addlegendentry{$\sigma_{\textrm{avg}}$}
\end{groupplot}

\end{tikzpicture}
    \caption{Standard deviation with respect to ramp-up times, stroke gain, and pinching diameter. The deviation increases significantly for ramp-up times lower than \SI{0.5}{\second}. The standard deviations stay consistent for ramp-up times larger than \SI{0.5}{\second} up until continuous operation (illustrated as dashed line). Small stroke gains show small deviations, while high stroke gain result in large differences between the deviation in x (along long table edge) and y (along short table edge) direction. The differences in deviations for ball pinch are small. High pinching diameters show slightly increased deviations.}
    \label{fig:systemparameters}
\end{figure}
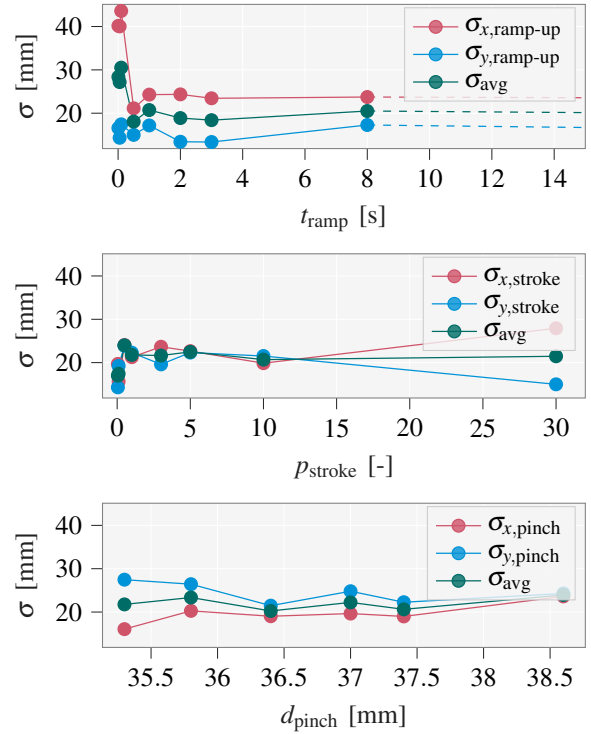
\label{sec:accuracy}
This section provides a detailed evaluation of the defined objectives of Section~\ref{sec:design}.
We first introduce the experimental setup, followed by an investigation of the ball launcher's accuracy for different hardware settings and distances. We close this section by comparing AIMY's capability to generate ball speed and spin in comparison to professional human table tennis players.
\begin{figure*}[t]
    \centering
    \includegraphics[width=\textwidth]{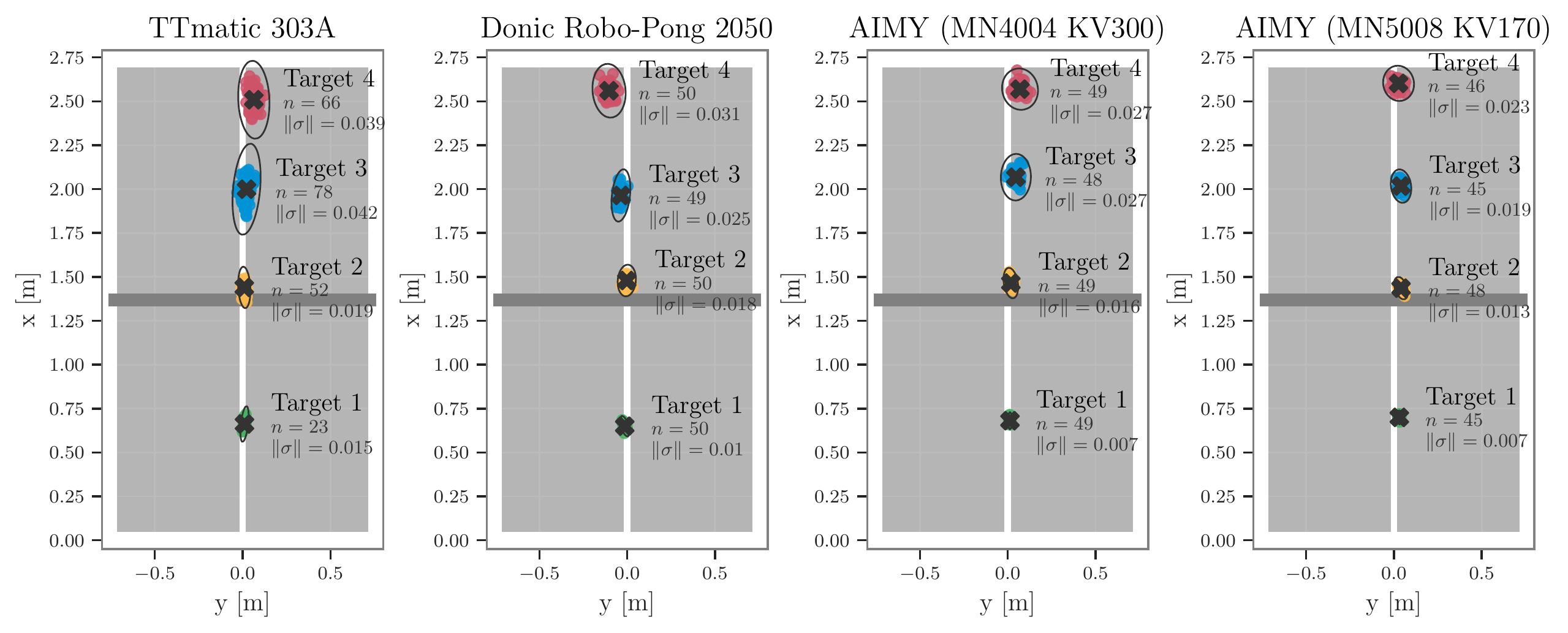}
    \caption{Distance accuracy experiment for each ball launcher with medium altitude angle. The standard deviation for different target positions for each ball launcher is evaluated. Each measurement point represents an approximated landing point of a ball trajectory on the table and is colored according to the affiliation to each distance measuring series. Additional information is given for each measuring series, like the number of measurement samples, the mean and confidence ellipses with 4 times the standard deviation, and the norm of the respective standard deviation $\lVert \mathbf{\sigma} \rVert$ are indicated for each measuring series.}
    \label{fig:launcher_comparison}
\end{figure*}

\subsection{Experimental Setup}
\label{sec:setup}
For the evaluation, four optical high-speed cameras (Prosilica Gigabit GE640C) record the ball trajectories at 200 frames per second. The tracking system is presented in~\cite{gomez2019}. 
For assessing the ball launcher's accuracy, we evaluate the standard deviation of the landing points of the respective ball trajectories. 
Due to the scatter of measurements and the discretization of the true trajectory, two polynomials are fitted before and after the sample with the lowest height. 
The intersection of the polynomials provides a better approximation of the true landing point than the data point with the lowest height. First-order polynomials are sufficient for this purpose and show the most reliable results. 
Before evaluation, a filter removes erroneous measurement samples of the recorded ball trajectories, and we inspected the ball trajectories manually for plausibility.

\subsection{Evaluation of System Parameters}
\label{sec:eval_sys}
Besides the control parameters, we identified three system parameters (the ramp-up time of the motors, the stroke gain of the ball feeding unit and ball pinching in the launch unit). We evaluate the influence of these parameters on the reproducibility of launched ball trajectories in the following. 
Figure~\ref{fig:systemparameters} depicts the accuracy evaluation of all three parameters. 
The ramp-up time is the time between launch initiation and actuation of the ball supply unit. 
During this time, the motors accelerate from standstill to the required speed. 
Figure~\ref{fig:systemparameters} shows that ramp-up times below 0.5~s have a negative influence on the accuracy of the ball launcher.
Measurements in continuous operation show comparable deviations of ramp-up times above \SI{2}{\second} and simultaneously provide a smoother operation of the motors.
The stroke gain influences the speed at which the balls are fed to the launch unit via the crank mechanism. With each control pass, the stroke gain increases the angle of the servo motor, causing steady insertion of the balls to the throwing wheels in the launch unit. The evaluation in Figure~\ref{fig:systemparameters} shows that small stroke gains result in smaller deviations, while increasing the total launch time. A stroke gain of 0.05, e.g., requires \SI{4.21}{\second} from actuation until launch, while a stroke gain of 1.0 only takes \SI{1.39}{\second}.
Ball pinching refers to the compression of the ball by the wheels during the launch procedure. The individual wheel units can be installed on the launch unit with different radial distances to the center. The smaller the distance, the higher the pressure generated on the ball during the launching process. The evaluation shows that increased pinching results in lower deviation of the landing points.

\subsection{High-fidelity Launching}
\label{sec:eval_fidelity}
To assess the fidelity of AIMY, the accuracy is evaluated in comparison with the commercially available ball launchers TTmatic 303A and Donic Robo-Pong 2050 for four different target distances. Since the characteristics of motors powering the throwing wheels have a major influence on accuracy, we selected and evaluated two different motors, the T-MOTOR Antigravity MN4004 KV300 and the T-MOTOR Antigravity MN5008 KV170 motors. Both motors have been installed with the same system parameters, control parameters, and experimental conditions. Launcher-specific differences of the custom-developed launcher to the commercial launchers due to different operation principles are reduced by the selection of comparable launch altitude angles and mechanical guides.
Figure~\ref{fig:launcher_comparison} shows that AIMY with both motors has a lower deviation of the landing points of Target 1, 2 and 4 than both commercial products. The TTmatic~303A showed an increased deviation on Target~3, which was caused by a parameter jump of the ball feed mechanism for low launch rates. The Robo-Pong 2050 shows a slighly smaller deviation on Target~3 than AIMY with MN4004 motors. The MN5008~motors have lower deviations on Target~2 to 4 than the MN4004 version, while MN4004 has higher accuracy on Target~1. However, the differences of both motors on Target~1 are not significant.  Similar results can be observed for the lowest and highest altitude settings in other measurement series. In conclusion, based on the experiments, it can be stated that AIMY has higher fidelity than the commercial products. In the presented measurement series, no outlier was detected or manually removed. Therefore, AIMY satisfies design objective~R2 regarding the accuracy, reliable operation and a low number of outliers.

\subsection{Suitability for Reinforcement Learning}
The sensor for the state of filling of the ball supply channel and the reservoir stirrer ensure a continuous and reliable supply of the launch unit. In the measurement series presented here, balls could be reliably shot remotely and at arbitrary times with delay times lower than \SI{500}{\milli\second} via the Python API. Based on the ramp-up time evaluation, we recommend setting a delay of three seconds in case of noncontinuous operation of the motors. In summary, AIMY sufficiently satisfies suitability for long-duration operation (R3) and fully satisfies remote controllability (R4).

\subsection{Comparison to Human Performance}
A maximum ball speed of \SI{15.4}{\metre\per\second} generated by AIMY is measured, which is slightly lower than the average maximum force serves with \SI{16.8}{\metre\per\second} of national team level players measured by Lee et al. \cite{lee2019}. 
AIMY generates top spins up to \SI{192.0}{\per\second}, measured by observing the change of orientation of marked table tennis balls with 960 frames per second recordings. 
This value significantly exceeds the average of \SI{113.9}{\per\second} measured for human professional players. 
Ball speeds and ball spins have been recorded independently of each other.

\section{Target Shooting}
\label{sec:target}
AIMY enables precise adjustment of training conditions for learning algorithms. To be able to set the ball trajectories to meet the demands of controllable training scenarios, we developed an algorithm which selects suitable control parameters based on the desired state parameters, e.g., the landing point, initial rotational yaw, roll of the ball, angle of impact of the landing point, or rebound direction. The use of an analytical model introduces approximation errors of non-linear effects such as inelastic and rolling collision dynamics at the rebound and friction and slip in the launch unit. Therefore, we decided to use a feed-forward neural network for the target shooting algorithm. 
\begin{table}[b]
\centering
\caption{Characteristics of the training data set}
\label{tab:dataset}
\begin{tabular}{@{}ccccc@{}}
\toprule
\textbf{\#} & \begin{tabular}[c]{@{}c@{}}\textbf{Number}\\ \textbf{samples}\end{tabular} & \textbf{\begin{tabular}[c]{@{}c@{}}Spin\\ magnitude\end{tabular}} & \textbf{\begin{tabular}[c]{@{}c@{}}Wheel \\ speeds\end{tabular}} & \textbf{\begin{tabular}[c]{@{}c@{}}Altitude \\ angle\end{tabular}} \\ \midrule
1 & 415 & \begin{tabular}[c]{@{}c@{}}none to \\ low\end{tabular} & \begin{tabular}[c]{@{}c@{}}all wheels \\ same\end{tabular} & various \\ \midrule
2 & 64 & high & high speeds & various \\ \midrule
3 & 364 & \begin{tabular}[c]{@{}c@{}}low to \\ medium\end{tabular} & low speeds & various \\ \midrule
4 & 1103 & \begin{tabular}[c]{@{}c@{}}low to \\ high\end{tabular} & various & high \\ \midrule
5 & 1385 & \begin{tabular}[c]{@{}c@{}}low to \\ medium\end{tabular} & various & medium \\ \midrule
6 & 430 & low & various & low \\ \midrule
\textbf{Total} & \textbf{3761} &  &  &  \\ \bottomrule
\end{tabular}
\end{table}

\subsection{Training Data}
A balanced and rich data set has a decisive influence on whether suitable control parameters can be found for the desired state parameters. A grid selects a balanced set of control parameters for the training data set of the target shooting algorithm. Table~\ref{tab:dataset} describes the composition of the data used for target shooting. The data is biased towards hitting the opponent's half of the table.
Therefore, higher velocities and higher angles occur more often. The data set also includes trajectories that do not bounce on the table. Of the 3761 trajectories provided, 3250 have a landing point on the table. Our project web page provides the used data set, the trained model for target shooting along with evaluation scripts.

\subsection{Evaluation}
\label{sec:target_eval}
\begin{figure}[t]
    \centering
    \includegraphics[width=0.41\textwidth]{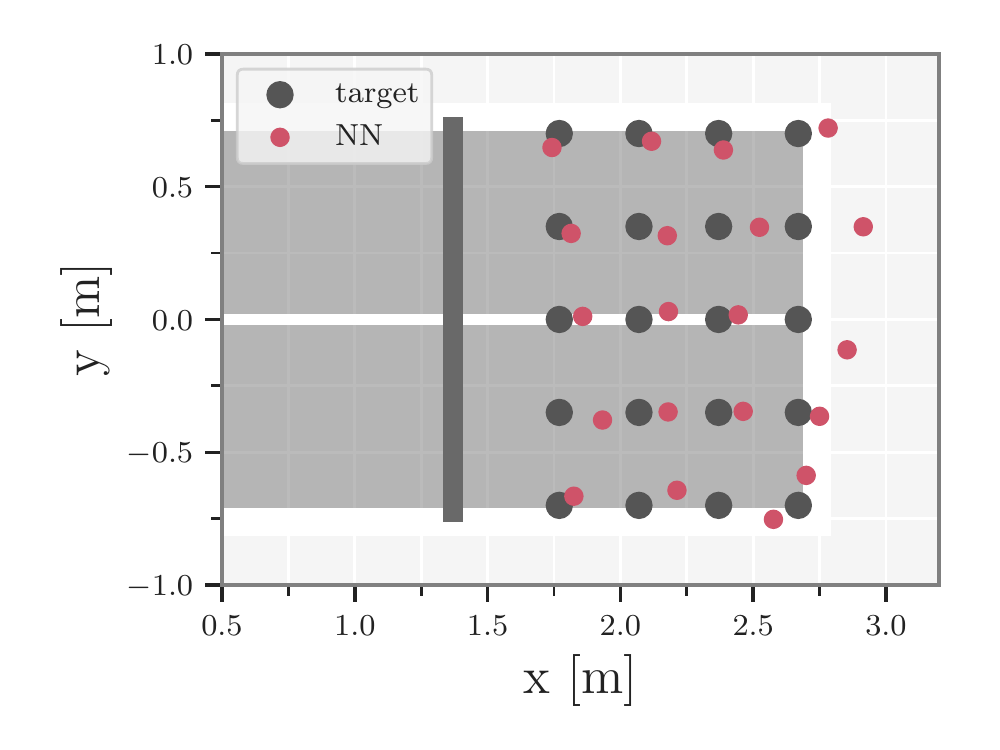}
    \caption{Target shooting experiment. Grey markers represent the targets. Red markers indicate landing positions of ball trajectories generated with the neural network approach. While the landing points roughly match the target locations, deviations are visible.}
    \label{fig:targetshooting}
\end{figure}

The model consists of a feedforward, fully connected neural network with three hidden layers. Input to the model is the desired landing position in Cartesian coordinates. 
The model outputs the set of control parameters. 
For evaluation, we selected a grid of 20 points on the opposite half of the table. Figure~\ref{fig:targetshooting} illustrates the result.
The obtained landing points indicate that the model can roughly hit desired landing points.

\section{Conclusion}

In this work, we provide an open-source table tennis ball launcher capable of generating versatile ball trajectories in a reproducible and controlled manner. The design satisfies the requirements to enable long-duration training of robot table tennis. AIMY has higher accuracy than commercially available products and additionally features remote APIs, crucial for embedding in an RL setting. 
In a first learning attempt, we showed that AIMY shoots balls close to a desired landing position on the table. The limitations of this work include a missing comparison of AIMY with latest commercial products like the Butterfly AMICUS Prime and the precision of the proposed target shooting approach. More sophisticated approaches using RL or incorporate analytical knowledge of the ball launcher and the ball dynamics could fully utilize AIMY's accuracy.
In future work, we will investigate how to return and generate different kinds of spin with a real robot.
We deeply hope that AIMY will contribute to increased performance of table tennis robots in the future. 
Videos of the experiments, as well as construction details, can be found on the project webpage.

\section*{Acknowledgements}
We thank Vincent Berenz and Felix Widmaier for their support developing the software.


\addtolength{\textheight}{-10.5cm}   


\bibliographystyle{IEEEtran}
\bibliography{reference}

\begin{thebibliography}{10}
\providecommand{\url}[1]{#1}
\csname url@rmstyle\endcsname
\providecommand{\newblock}{\relax}
\providecommand{\bibinfo}[2]{#2}
\providecommand\BIBentrySTDinterwordspacing{\spaceskip=0pt\relax}
\providecommand\BIBentryALTinterwordstretchfactor{4}
\providecommand\BIBentryALTinterwordspacing{\spaceskip=\fontdimen2\font plus
\BIBentryALTinterwordstretchfactor\fontdimen3\font minus
  \fontdimen4\font\relax}
\providecommand\BIBforeignlanguage[2]{{%
\expandafter\ifx\csname l@#1\endcsname\relax
\typeout{** WARNING: IEEEtran.bst: No hyphenation pattern has been}%
\typeout{** loaded for the language `#1'. Using the pattern for}%
\typeout{** the default language instead.}%
\else
\language=\csname l@#1\endcsname
\fi
#2}}

\bibitem{tebbe2021}
J.~Tebbe, L.~Krauch, Y.~Gao, and A.~Zell, ``Sample-efficient reinforcement
  learning in robotic table tennis,'' in \emph{2021 IEEE International
  Conference on Robotics and Automation (ICRA)}, 2021, pp. 4171--4178.

\bibitem{luo2021}
L.~Yang, H.~Zhang, X.~Zhu, and X.~Sheng, ``Ball motion control in the table
  tennis robot system using time-series deep reinforcement learning,''
  \emph{IEEE Access}, vol.~9, pp. 99\,816--99\,827, 2021.

\bibitem{buechler2022}
D.~Büchler, S.~Guist, R.~Calandra, V.~Berenz, B.~Schölkopf, and J.~Peters,
  ``Learning to play table tennis from scratch using muscular robots,''
  \emph{IEEE Transactions on Robotics}, pp. 1--11, 2022.

\bibitem{abueyruwan2022}
\BIBentryALTinterwordspacing
S.~Abeyruwan, L.~Graesser, D.~B. D'Ambrosio, A.~Singh, A.~Shankar, A.~Bewley,
  and P.~R. Sanketi, ``i-sim2real: Reinforcement learning of robotic policies
  in tight human-robot interaction loops,'' 2022. [Online]. Available:
  \url{https://arxiv.org/abs/2207.06572}
\BIBentrySTDinterwordspacing

\bibitem{lee2019}
\BIBentryALTinterwordspacing
M.~J.~C. Lee, H.~Ozaki, and W.~X. Goh, ``Speed and spin differences between the
  old celluloid versus new plastic table tennis balls and the effect on the
  kinematic responses of elite versus sub-elite players,'' \emph{International
  Journal of Racket Sports Science}, vol.~1, 6 2019. [Online]. Available:
  \url{http://hdl.handle.net/10481/57324}
\BIBentrySTDinterwordspacing

\bibitem{Matsushima2005}
M.~Matsushima, T.~Hashimoto, M.~Takeuchi, and F.~Miyazaki, ``A learning
  approach to robotic table tennis,'' \emph{IEEE Transactions on Robotics},
  vol.~21, no.~4, pp. 767--771, 2005.

\bibitem{huang2016jointly}
Y.~Huang, D.~B{\"u}chler, O.~Ko{\c{c}}, B.~Sch{\"o}lkopf, and J.~Peters,
  ``Jointly learning trajectory generation and hitting point prediction in
  robot table tennis,'' in \emph{2016 IEEE-RAS 16th International Conference on
  Humanoid Robots (Humanoids)}.\hskip 1em plus 0.5em minus 0.4em\relax IEEE,
  2016, pp. 650--655.

\bibitem{malearning}
H.~Ma, D.~B{\"u}chler, B.~Sch{\"o}lkopf, and M.~Muehlebach, ``A learning-based
  iterative control framework for controlling a robot arm with pneumatic
  artificial muscles,'' in \emph{Robotics: Science and Systems}, 2022.

\bibitem{koc2018}
\BIBentryALTinterwordspacing
O.~Koç, G.~Maeda, and J.~Peters, ``Online optimal trajectory generation for
  robot table tennis,'' \emph{Robotics and Autonomous Systems}, vol. 105, pp.
  121--137, 2018. [Online]. Available:
  \url{https://www.sciencedirect.com/science/article/pii/S0921889017306164}
\BIBentrySTDinterwordspacing

\bibitem{gomez2020adaptation}
S.~Gomez-Gonzalez, G.~Neumann, B.~Sch{\"o}lkopf, and J.~Peters, ``Adaptation
  and robust learning of probabilistic movement primitives,'' \emph{IEEE
  Transactions on Robotics}, vol.~36, no.~2, pp. 366--379, 2020.

\bibitem{muelling2013}
K.~Mülling, J.~Kober, O.~Kroemer, and J.~Peters, ``Learning to select and
  generalize striking movements in robot table tennis,'' \emph{The
  International Journal of Robotics Research}, vol.~32, no.~3, pp. 263--279,
  2013.

\bibitem{yang2021}
L.~Yang, H.~Zhang, X.~Zhu, and X.~Sheng, ``Ball motion control in the table
  tennis robot system using time-series deep reinforcement learning,''
  \emph{IEEE Access}, vol.~9, pp. 99\,816--99\,827, 2021.

\bibitem{abeyruwan2022}
S.~Abeyruwan, L.~Graesser, D.~B. D'Ambrosio, A.~Singh, A.~Shankar, A.~Bewley,
  and P.~R. Sanketi, ``i-sim2real: Reinforcement learning of robotic policies
  in tight human-robot interaction loops,'' \emph{arXiv preprint
  arXiv:2207.06572}, 2022.

\bibitem{ttmatic303}
``{TTmatic 303 A},''
  \url{http://www.megaspin.net/download/ttmatic/2007_catalog.pdf}, accessed:
  2022-08-21.

\bibitem{robopong3050xl}
``{Donic Newgby Robo-Pong 2050},''
  \url{https://www.donic.com/donic/roboter/3455/donic-newgy-robo-pong-2055-inkl.-72-donic-coach-p40-baelle?c=29},
  accessed: 2022-08-21.

\bibitem{powerpongomega}
``{Power Pong Omega},''
  \url{https://www.powerpong.org/collections/professional-robots/products/power-pong-omega-table-tennis-robot},
  accessed: 2022-08-21.

\bibitem{amicusprime}
``{AMICUS PRIME Roboter},'' \url{https://de.butterfly.tt/amicus-expert.html},
  accessed: 2022-08-21.

\bibitem{nemire1991}
\BIBentryALTinterwordspacing
K.~Nemire, M.~Goettsche, and B.~Bridgeman, ``Automated system for ball
  launching, visual occlusion, and data acquisition in a ball-hitting task,''
  \emph{Behavior Research Methods, Instruments, {\&} Computers}, vol.~23,
  no.~1, pp. 36--44, Mar 1991. [Online]. Available:
  \url{https://doi.org/10.3758/BF03203333}
\BIBentrySTDinterwordspacing

\bibitem{ponnusamy2006}
B.~Ponnusamy, W.~F. Yong, and Z.~Ahmad, ``A low cost automated table tennis
  launcher,'' \emph{ARPN Journal of Engineering and Applied Sciences}, vol.~10,
  no.~1, pp. 291--296, Jan 2006.

\bibitem{wuthrich2020}
M.~W{\"u}thrich, F.~Widmaier, F.~Grimminger, J.~Akpo, S.~Joshi, V.~Agrawal,
  B.~Hammoud, M.~Khadiv, M.~Bogdanovic, V.~Berenz, \emph{et~al.}, ``Trifinger:
  An open-source robot for learning dexterity,'' \emph{arXiv preprint
  arXiv:2008.03596}, 2020.

\bibitem{grimminger2020}
F.~Grimminger, A.~Meduri, M.~Khadiv, J.~Viereck, M.~W{\"u}thrich, M.~Naveau,
  V.~Berenz, S.~Heim, F.~Widmaier, T.~Flayols, \emph{et~al.}, ``An open
  torque-controlled modular robot architecture for legged locomotion
  research,'' \emph{IEEE Robotics and Automation Letters}, vol.~5, no.~2, pp.
  3650--3657, 2020.

\bibitem{gomez2019}
\BIBentryALTinterwordspacing
S.~Gomez-Gonzalez, Y.~Nemmour, B.~Schölkopf, and J.~Peters, ``Reliable
  real-time ball tracking for robot table tennis,'' \emph{Robotics}, vol.~8,
  no.~4, 2019. [Online]. Available: \url{https://www.mdpi.com/2218-6581/8/4/90}
\BIBentrySTDinterwordspacing

\end{thebibliography}


\addtolength{\textheight}{+10.5cm}   

\newpage
\onecolumn
\raggedbottom
\section*{APPENDIX}

\subsection{Technical Specifications}
Table~\ref{tab:specifications} provides a brief overview of the specifications of AIMY.

\begin{table}[ht]
\centering
\caption{Technical Specifications}
\label{tab:specifications}
\begin{tabular}{@{}lcl@{}}
\toprule
\textbf{Specification} & \textbf{Value} & \textbf{Remarks} \\ \midrule
Maximum Speed & \SI{15.4}{\metre\per\second} & Average speed in first \SI{0.5}{\second} \\
Maximum Spin & \SI{192.0}{\per\second} & Initial spin \\ 
Altitude Angle & \ang{6.4} -- \ang{37.1} & \\
Azimuthal Angle & \ang{-15.8} -- \ang{15.6}& \\
Supply Voltage & \SI{230}{\volt} (\SI{50}{\hertz})& \\
\midrule
Python API & \checkmark & \\
C++ API & (\checkmark) & Not extensively tested \\
Ethernet & \checkmark & \\
Wi-Fi & \checkmark & \\
\bottomrule
\end{tabular}
\end{table}

\subsection{Third Party Hardware}
Table~\ref{tab:thirdparty} lists third party hardware selected for AIMY. In Section~\ref{sec:eval_fidelity}, the evaluation assessed two different motors, both being listed. We recommend the T-MOTOR Antigravity MN5008 KV170, as it provides higher torque in the lower turning speed region. The ball launcher mainly operates in lower turning speed regions, which results in higher accuracy in comparison to the T-MOTOR Antigravity MN4004 KV300.

\begin{table}[H]
\centering
\caption{}
\label{tab:thirdparty}
\begin{tabular}{@{}lll@{}}
\toprule
\textbf{Component} & \textbf{Manufacturer} & \textbf{Model} \\ \midrule
Motor Wheels & T-MOTOR & Antigravity MN4004 KV300 \\
Motor Wheels & T-MOTOR & Antigravity MN5008 KV170 \\
ESC & T-MOTOR & AIR 40A \\
\begin{tabular}[c]{@{}l@{}}Servomotor Azimuthal Orientation\end{tabular} & Hitec & D954SW \\
\begin{tabular}[c]{@{}l@{}}Servomotor Altitude Orientation\end{tabular} & Actuoinx & L12-30-210-6-R \\
\begin{tabular}[c]{@{}l@{}}Servomotor Reservoir Stirrer\end{tabular} & Pololu & SpringRC SM-S4303R \\
\begin{tabular}[c]{@{}l@{}}Servomotor Ball Feeding Unit\end{tabular} & Hitec & HSB-9381TH \\
\begin{tabular}[c]{@{}l@{}}Optical Sensor Ball Feeding Unit\end{tabular} & Pepperl+Fuchs & GLV18-8-400-S \\
Control Unit & \begin{tabular}[c]{@{}l@{}}Raspberry Pi Foundation\end{tabular} & Raspberry Pi 4 Model B \\
Motor Driver Wheels & Xinabox & OC05 \\
24 V Power Supply Unit & Mean Well & HDR-150-24 \\
5 V Power Supply Unit & Mean Well & MDR-40-5 \\
Azimuthal Bearing & Franke & LEL1\_5\_0070 \\
Interface & \begin{tabular}[c]{@{}l@{}}Raspberry Pi Foundation\end{tabular} & \begin{tabular}[c]{@{}l@{}}Raspberry Pi Touch \\ Display (7-inch)\end{tabular} \\
Throw wheel & Butterfly & AMICUS Ball Throw Wheel \\
Housing & item & various \\ \bottomrule
\end{tabular}
\end{table}

\subsection{Experimental Setup}
In addition to the description of the experimental setup in Section~\ref{sec:setup}, we provide further details on the setup, and the approach of processing and collecting data for evaluation. After recording and before evaluation, we transform the data points to a common coordinate system and preprocess the trajectories according to the following procedure.

\begin{itemize}
    \item \textbf{Measurement System:} The accuracy of the ball tracking system has been evaluated beforehand. We measured a deviation of \SI{0.024}{\metre} in x-direction, \SI{0.011}{\metre} in y-direction and \SI{0.001}{\metre} in z-direction per meter within the relevant recording region.
    \item \textbf{Table Tennis Launcher:} For high-fidelity evaluation, we assess our custom-designed ball launcher in two different configurations. The first configuration is equipped with the T-MOTOR MN4004 KV300 and the second configuration equipped with the T-MOTOR MN4004 KV300 for accelerating balls in the desired direction. In addition to the motors, the structure of the launch unit was made mechanically stiffer in the second configuration, which may have an impact on accuracy in addition to motor differences. For system parameter evaluation, the second configuration is used. In the current version of AIMY, there are no sensors installed for measuring turning speeds of the motors and providing closed-loop control of turning speeds. Individual motor differences in motor speeds for the same actuation are compensated with measured characteristic motor curves for each motor. The characteristic motor curve maps actuation and resulting motor speeds and interpolates between each measurement point.  Figure~\ref{fig:characteristics} shows the characteristic motor curves. Figure~\ref{fig:orientationactors} shows the measured actuation to angle change relationship of the orientation actuators.
    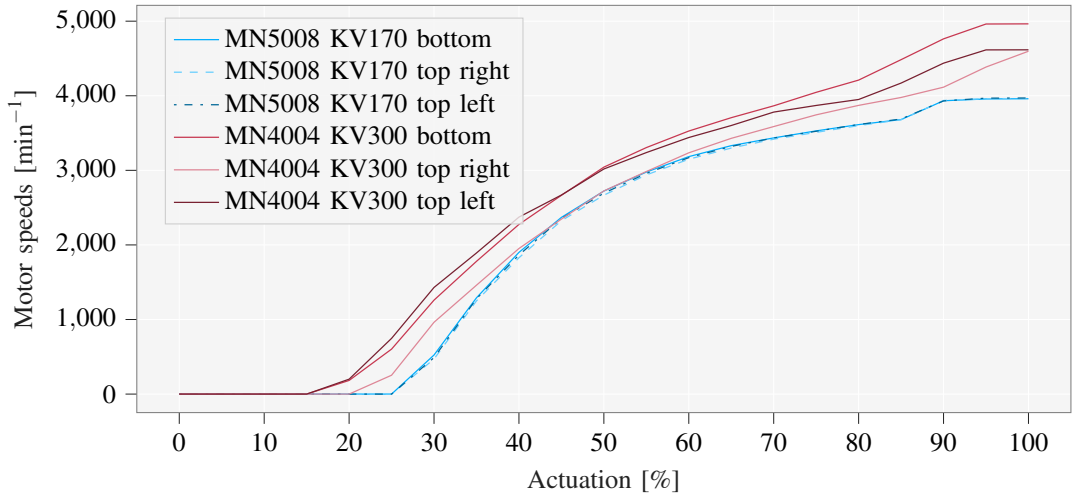
\begin{figure}[ht]
        \centering
\begin{tikzpicture}

\definecolor{black34}{RGB}{34,34,34}
\definecolor{brown1203349}{RGB}{120,33,49}
\definecolor{deepskyblue1174254}{RGB}{1,174,254}
\definecolor{gray}{RGB}{128,128,128}
\definecolor{indianred1995681}{RGB}{199,56,81}
\definecolor{lightgray204}{RGB}{204,204,204}
\definecolor{lightskyblue103206254}{RGB}{103,206,254}
\definecolor{palevioletred222135151}{RGB}{222,135,151}
\definecolor{teal1104152}{RGB}{1,104,152}
\definecolor{whitesmoke}{RGB}{245,245,245}

\begin{axis}[
axis background/.style={fill=whitesmoke},
axis line style={gray},
height=7.0cm,
legend cell align={left},
legend style={
  fill opacity=0.8,
  draw opacity=1,
  text opacity=1,
  at={(0.03,0.97)},
  anchor=north west,
  draw=lightgray204,
  fill=whitesmoke
},
tick align=outside,
tick pos=left,
width=14.0cm,
x grid style={white},
xlabel=\textcolor{black34}{Actuation [\%]},
xmajorgrids,
xmin=-5, xmax=105,
xtick style={color=black34},
y grid style={white},
ylabel=\textcolor{black34}{Motor speeds [min\(\displaystyle ^{-1}\)]},
ymajorgrids,
ymin=-248.2, ymax=5212.2,
ytick style={color=black34}
]
\addplot [line width=0.48pt, deepskyblue1174254]
table {%
0 0
5 0
10 0
15 0
20 0
25 0
30 524
35 1291
40 1900
45 2369
50 2722
55 2977
60 3184
65 3331
70 3435
75 3529
80 3613
85 3681
90 3933
95 3957
100 3960
};
\addlegendentry{MN5008 KV170 bottom}
\addplot [line width=0.48pt, lightskyblue103206254, dashed]
table {%
0 0
5 0
10 0
15 0
20 0
25 0
30 470
35 1250
40 1828
45 2325
50 2667
55 2939
60 3149
65 3299
70 3418
75 3509
80 3600
85 3678
90 3937
95 3961
100 3963
};
\addlegendentry{MN5008 KV170 top right}
\addplot [line width=0.48pt, teal1104152, dash pattern=on 1pt off 3pt on 3pt off 3pt]
table {%
0 0
5 0
10 0
15 0
20 0
25 0
30 489
35 1278
40 1865
45 2354
50 2694
55 2959
60 3170
65 3321
70 3434
75 3524
80 3615
85 3688
90 3931
95 3968
100 3970
};
\addlegendentry{MN5008 KV170 top left}
\addplot [line width=0.48pt, indianred1995681]
table {%
0 0
5 0
10 0
15 0
20 183
25 603
30 1260
35 1780
40 2275
45 2665
50 3043
55 3305
60 3527
65 3706
70 3865
75 4046
80 4210
85 4484
90 4763
95 4962
100 4964
};
\addlegendentry{MN4004 KV300 bottom}
\addplot [line width=0.48pt, palevioletred222135151]
table {%
0 0
5 0
10 0
15 0
20 0
25 253
30 963
35 1460
40 1952
45 2346
50 2724
55 2984
60 3235
65 3429
70 3587
75 3744
80 3871
85 3977
90 4115
95 4385
100 4597
};
\addlegendentry{MN4004 KV300 top right}
\addplot [line width=0.48pt, brown1203349]
table {%
0 0
5 0
10 0
15 0
20 200
25 747
30 1430
35 1892
40 2373
45 2669
50 3018
55 3238
60 3440
65 3601
70 3781
75 3870
80 3950
85 4170
90 4437
95 4615
100 4616
};
\addlegendentry{MN4004 KV300 top left}
\end{axis}

\end{tikzpicture}
        \caption{Motor characteristic curves of each evaluated motors with the available actuation as input and the turning speed as output. The MN5008 show similar curves for each position, while the curves of the MN4004 have significant differences in behavior.}
        \label{fig:characteristics}
    \end{figure}
    \begin{figure}[H]
        \centering
        \includegraphics[width=0.5\textwidth]{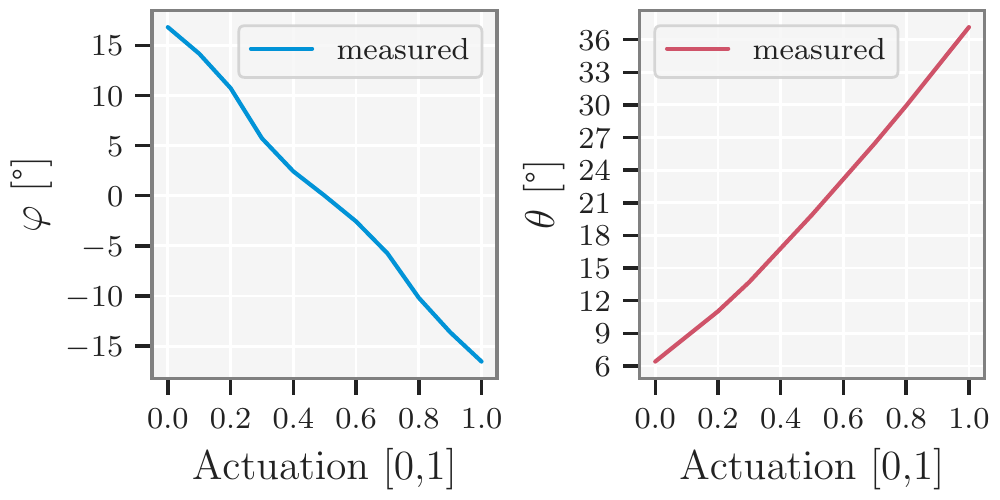}
        \caption{Characteristics of the orientation actuators for different actuation signals.}
        \label{fig:orientationactors}
    \end{figure}
    
    \item \textbf{Setup:} For reducing discrepancies between the different structures of the two configurations of the custom designed ball launcher and the TTmatic~303A, similar altitude angles are chosen for evaluation and specified with the terms "low", "medium" and "high". The ball launchers are positioned directly at the front-end of the table. The TTmatic~303A is not capable of launching the small distance to Target~1 with automatic ball feed. Therefore, the distance of the TTmatic~303A to the front-end of the table had to be increased by approximately \SI{30}{\centi\meter}. For target shooting, the data set has been recorded with AIMY in the second configuration with a distance of \SI{80}{\centi\meter} to the front-end of the table. With increased distance higher speeds and spins can be utilized, however, with a compromise on accuracy. 
\end{itemize}

The provided data sets have been processed before the -- in Section~\ref{sec:accuracy} described -- bounce detection is applied. The processing includes several filter processes addressing either faulty samples or faulty trajectories:
\begin{itemize}
    \item A time jump filter removes a trajectory from the data set, if there exists a time gap between two consecutive samples exceeding \SI{0.5}{\second}. 
    \item A rebound filter removes a trajectory from the data set, if there is no rebound on the table. However, trajectories closely missing the table edges provide important information for target shooting learning. Therefore, the data sets have been exported with rebound filter applied and without.
    \item The position jump filter removes samples with position deviations of more than \SI{5}{\centi\meter} between one sample and the neighborhood of 15 closest samples. This process proved to remove most outliers caused by the ball tracking system.
    \item The region filter removes all samples outside the table area, with some relaxation in x- and y-direction.
\end{itemize}

Additionally, all samples have been transformed to fit the defined initial coordinate system, with x increasing along the long edge of the table and y along the short edge of the table. We set the origin at the frontend of the table in the middle of the table. The time stamps are converted from nanoseconds to seconds. In addition to jump detection, we inspected all trajectories visually for rough outliers to ensure plausible training data and evaluation results.

\subsection{Evaluation of Orientation Jumps}
\begin{figure}[H]
    \centering
    \includegraphics[width=0.4\textwidth]{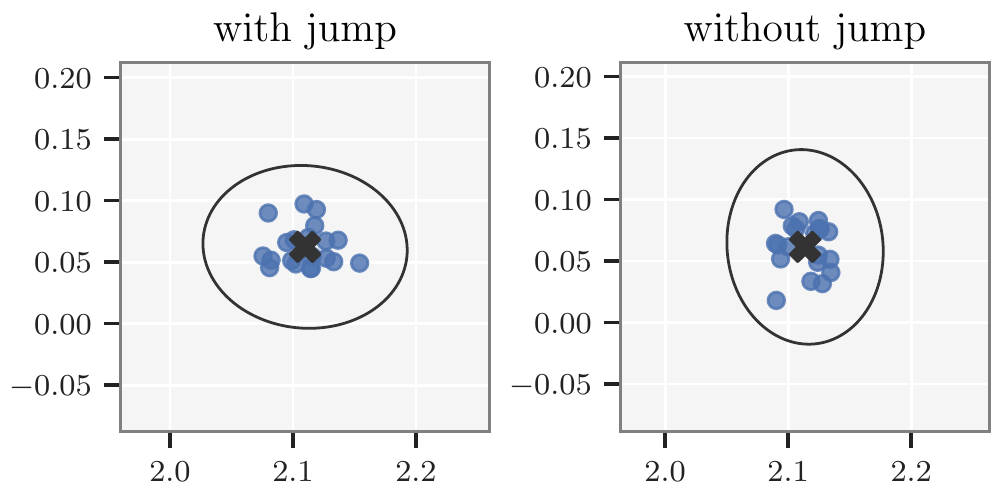}
    \caption{Orientation jump experiment. Each measurement point represents a landing point of a ball trajectory. The dimension of the confidence ellipse is roughly equal between the measurements with and without jump. However, the orientation of the larger deviation switched between the measurement series.}
    \label{fig:appendix_jump}
\end{figure}

Additionally to the system parameters assessed in Section~\ref{sec:accuracy}, we evaluated the influence of orientation jumps on the ball trajectories. Orientation jumps are significant changes in altitude and azimuthal angles of the launching tube . 
For evaluation, we recorded ball trajectories in which the orientation was not changed beforehand and launches in which an orientation jump was made beforehand. An altitude angle of \ang{19.9} and an azimuthal angle of 0° define the default launching position of the orientation actuators. Before each shot, the table tennis launcher moves into the neutral position from a displaced position with an altitude angle of \ang{6.4} and an azimuthal angle of \ang{-15.8}. 
Table~\ref{tab:jump} and Figure~\ref{fig:appendix_jump} provide the results. The figure shows no significant deviations of the landing points for trajectories recorded with orientation jumps in comparison to trajectories without orientation jumps. 
\begin{table}[H]
\centering
\caption{Investigation on the influence of orientation jumps on the landing points accuarcy}
\begin{tabular}{@{}cccc@{}}
\toprule
\begin{tabular}[c]{@{}c@{}}Orientation \\ jump\end{tabular} & \begin{tabular}[c]{@{}c@{}}Number\\ samples [-]\end{tabular} & $\sigma_x$ [mm]& $\sigma_y$ [mm]\\ \midrule
no & 20 & \textbf{0.0155} & 0.0193 \\
yes & 20 & 0.0195 & \textbf{0.0176} \\ \bottomrule
\end{tabular}
\label{tab:jump}
\end{table}

\subsection{Evaluation of Stroke Gain}
\begin{figure}[H]
    \centering
    \includegraphics[width=0.8\textwidth]{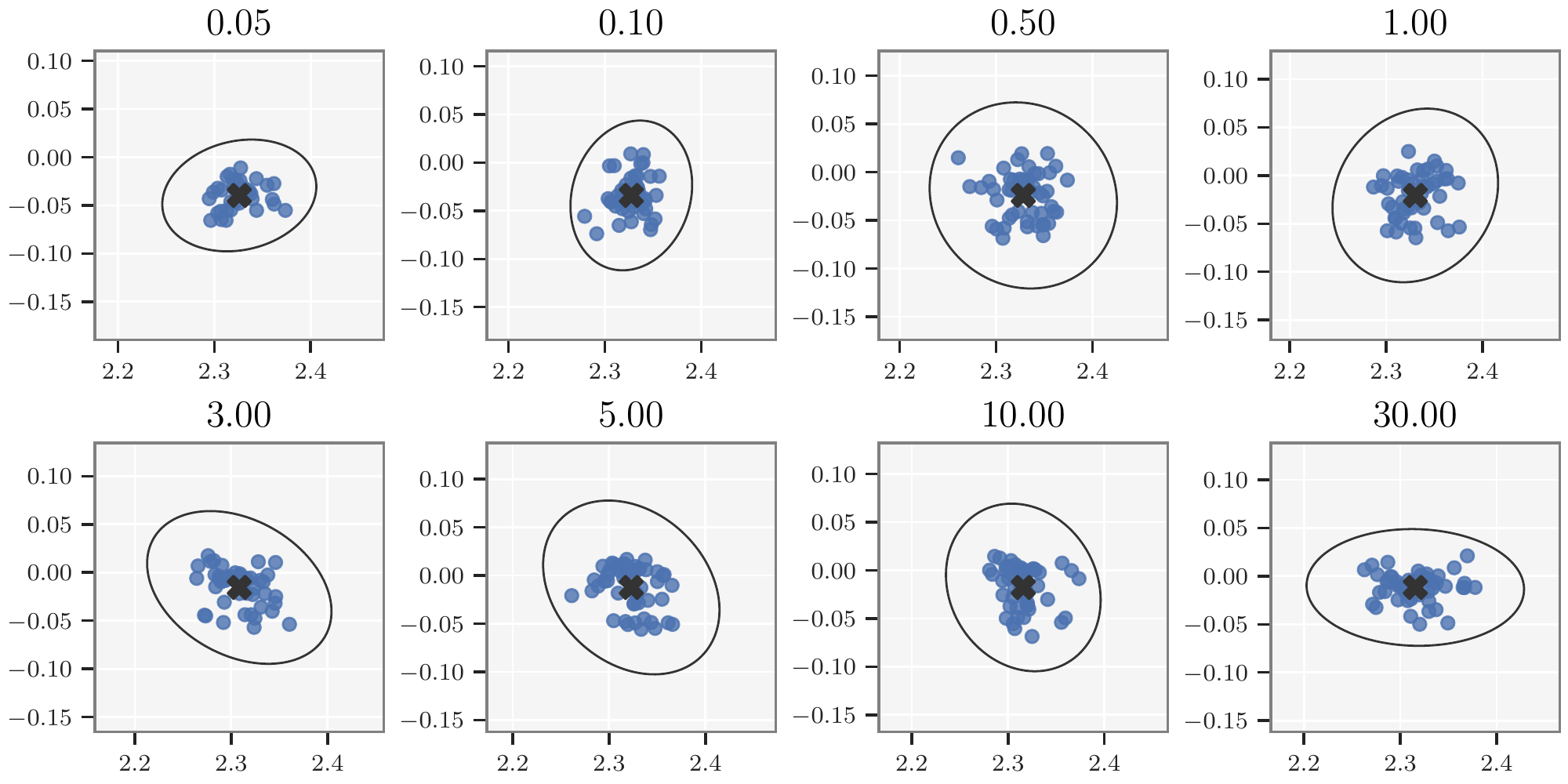}
    \caption{Stroke gain experiment. Landing points are illustrated for different stroke gain values along the confidence ellipses of three times the standard deviation, and the mean value given as a cross. One can observe only minor differences of the deviations between each stroke gain values.}
    \label{fig:appendix_stroke}
\end{figure}

Table~\ref{tab:strokegain} lists the standard deviation values in each direction of the table plane of Figure~\ref{fig:systemparameters} for different stroke gain values. Stroke gain is briefly defined in Section~\ref{sec:eval_sys}. Figure~\ref{fig:appendix_stroke} further illustrates the deviation of the landing points.

\begin{table}[H]
\centering
\caption{Investigation on the influence of stroke gain on the landing points accuarcy}
\label{tab:strokegain}
\begin{tabular}{@{}cccccc@{}}
\toprule
\begin{tabular}[c]{@{}c@{}}Stroke \\ gain [-]\end{tabular} & \begin{tabular}[c]{@{}c@{}}Number\\ samples\end{tabular} & \multicolumn{1}{c}{$\sigma_x$ [mm]} & \multicolumn{1}{c}{$\sigma_y$ [mm]} & \multicolumn{1}{c}{$A_\sigma$ [mm$^2$]} & \begin{tabular}[c]{@{}c@{}}Launch \\ time [s]\end{tabular} \\ \midrule
0.05 & 35 & 0.0197 & \textbf{0.0143} & \textbf{0.00088} & 8.83 \\
0.10 & 50 & \textbf{0.0156} & 0.0192 & 0.00095 & 4.21\\
0.50 & 50 & 0.0241 & 0.0239 & 0.00181 & 1.60\\
1.00 & 49 & 0.0213 & 0.0223 & 0.00149 & 1.39\\
3.00 & 46 & 0.0237 & 0.0196 & 0.00146 & 0.81\\
5.00 & 50 & 0.0227 & 0.0223 & 0.00159 & 0.61\\
10.00 & 50 & 0.0199 & 0.0215 & 0.00134 & 0.67\\
30.00 & 50 & 0.0279 & 0.0150 & 0.00132 & 0.62\\ \bottomrule
\end{tabular}
\end{table}

\subsection{Evaluation of Ball Pinching}
\begin{figure}[H]
    \centering
    \includegraphics[width=0.6\textwidth]{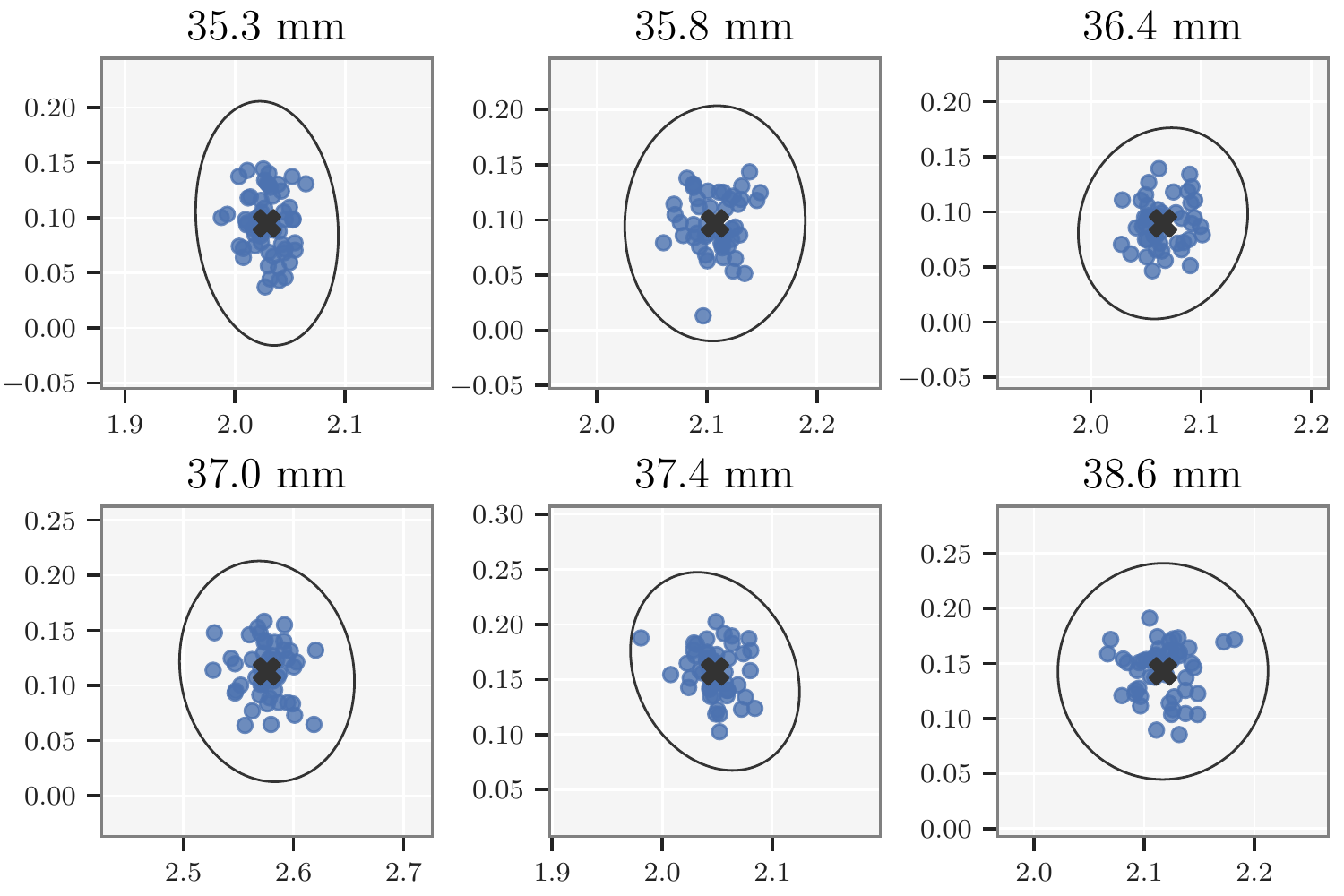}
    \caption{Ball pinching experiments. Landing points are illustrated for different ball pinching values along confidence ellipses of three times the standard deviation, and the mean value given as a cross. It can be seen that small and high pinching results in slightly increased deviations.}
    \label{fig:appendix_pinching}
\end{figure}

Table~\ref{tab:pinching} lists the standard deviation values in each direction of the table plane of Figure~\ref{fig:systemparameters} for different ball pinching values. Section~\ref{sec:eval_sys} gives a brief definition of ball pinching. Figure~\ref{fig:appendix_pinching} further illustrates the deviations of the landing points.

\begin{table}[H]
\centering
\caption{Investigation on the influence of ball pinching on the landing points accuarcy}
\begin{tabular}{@{}ccccc@{}}
\toprule
\begin{tabular}[c]{@{}c@{}}Pinching\\ diameter [mm]\end{tabular} & \begin{tabular}[c]{@{}c@{}}Number\\ samples [-]\end{tabular} & $\sigma_x$ [mm]& $\sigma_y$ [mm]& $A_\sigma$ [mm$^2$]\\ \midrule
35.3 & 51 & 0.03194 & 0.03062 & 0.00307 \\
35.8 & 48 & 0.01932 & 0.02658 & 0.00161 \\
36.4 & 49 & 0.01906 & \textbf{0.02171} & 0.00130 \\
37.0 & 49 & 0.01965 & 0.02480 & 0.00153 \\
37.4 & 49 & \textbf{0.01866} & 0.02208 & \textbf{0.00129} \\
38.6 & 49 & 0.02361 & 0.02428 & 0.00180 \\ \bottomrule
\end{tabular}
\label{tab:pinching}
\end{table}

\subsection{Evaluation of Motor Ramp-up}
\begin{figure}[H]
    \centering
    \includegraphics[width=0.8\textwidth]{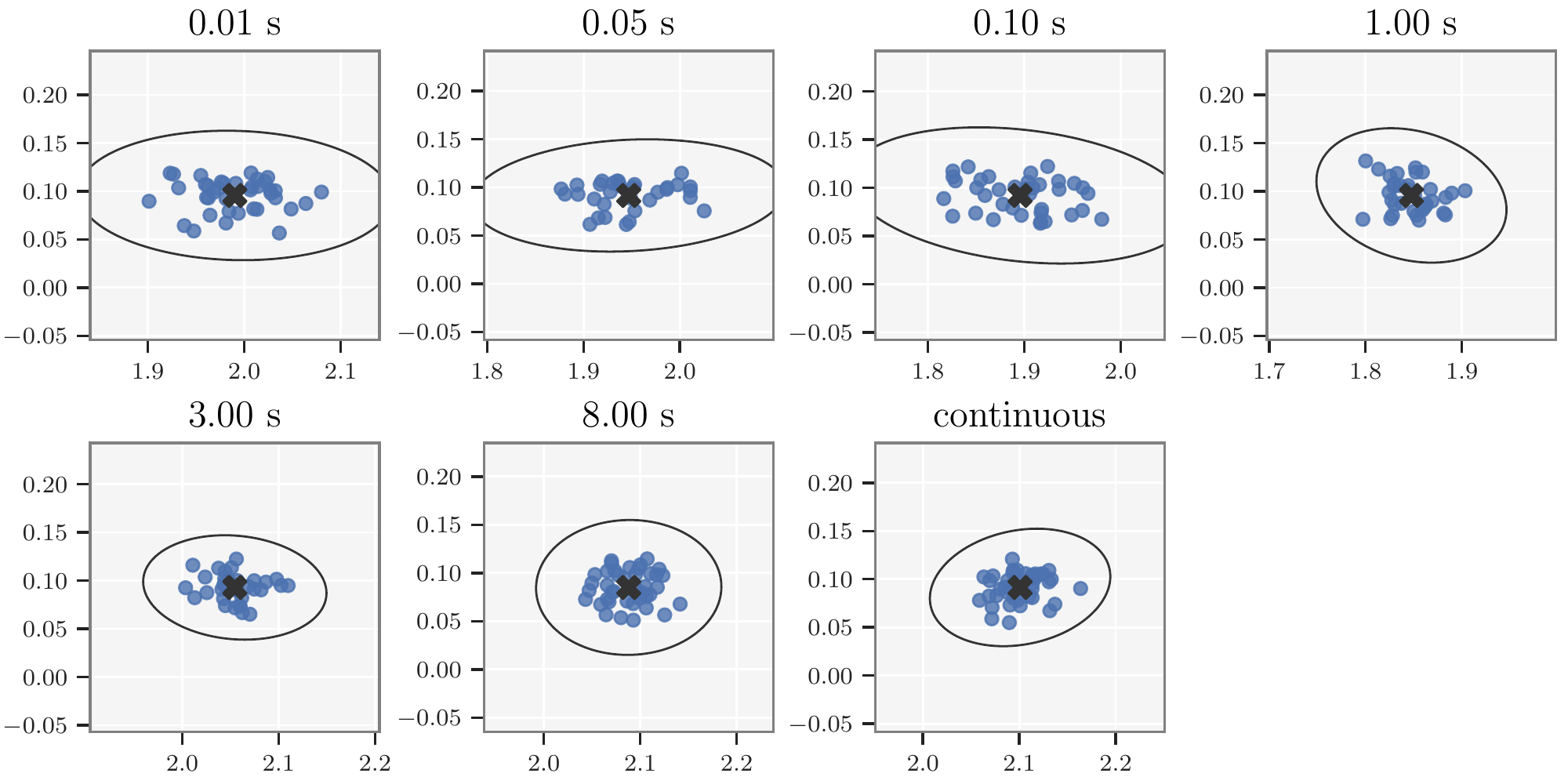}
    \caption{Ramp-up time experiments. Landing points are illustrated for different ramp-up times between actuation and launch of the ball along confidence ellipses of three times the standard deviation, and the mean value given as a cross. It can be seen that low ramp-up times result in significantly increased deviations in x-direction.}
    \label{fig:appendix_rampup}
\end{figure}

Table~\ref{tab:rampup} lists the standard deviation values in each direction of the table plane of Figure~\ref{fig:systemparameters} for different ramp-up times. Section~\ref{sec:eval_sys} gives a brief definition of the Ramp-up time. Figure~\ref{fig:appendix_rampup} further illustrates the deviations. The motors need a certain time to reach the requested speed. However, under 0.5 seconds of ramp-up time, the motor speeds are not yet converged at the time of launching. This noisy wheel velocities result in increased deviations of landing points.

\begin{table}[H]
\centering
\caption{Ramp-up times accuracy experiments}
\begin{tabular}{@{}cclll@{}}
\toprule
\begin{tabular}[c]{@{}c@{}}Ramp-up\\ time [s]\end{tabular} &
  \begin{tabular}[c]{@{}c@{}}Number\\ samples [-]\end{tabular} &
  \multicolumn{1}{c}{$\sigma_x$ [mm]} &
  \multicolumn{1}{c}{$\sigma_y$ [mm]} &
  \multicolumn{1}{c}{$A_\sigma$ [mm$^2$]} \\ \midrule
0.01      & 39 & 0.0402          & 0.0166          & 0.002096          \\
0.05      & 32 & 0.0400          & 0.0143          & 0.001797          \\
0.10       & 38 & 0.0436          & 0.0174          & 0.002383          \\
0.50       & 34 & \textbf{0.0211} & 0.0150          & 0.000994          \\
1.00       & 34 & 0.0243          & 0.0172          & 0.001313          \\
2.00       & 33 & 0.0243          & \textbf{0.0134} & 0.001023          \\
3.00       & 36 & 0.0234          & \textbf{0.0134} & \textbf{0.000985} \\
8.00       & 40 & 0.0247          & 0.0173          & 0.001342          \\
continuous & 40 & 0.0231          & 0.0150          & 0.001089          \\ \bottomrule
\end{tabular}
\label{tab:rampup}
\end{table}

\subsection{Evaluation of Versatility}
Table~\ref{tab:launcher_speed_comp} lists the experimentally determined maximum ball speed and topspin of the evaluated ball launchers. AIMY with the T--Motor N5008 KV170 motors is capable of generating the highest ball speeds and spins.

\begin{table}[th]
\centering
\caption{Comparison of maximum ball speeds and topspins of the evaluated ball launchers}
\label{tab:launcher_speed_comp}
\begin{tabular}{rrr}
\hline
 & \textbf{Max Speed} & \textbf{Max Topspin} \\ \hline
Professional Human (average) & \SI{16.8}{\metre\per\second} & \SI{113.9}{\per\second} \\
TTmatic 303A & \SI{9.8}{\metre\per\second} & \SI{64.0}{\per\second} \\
Donic Robo-Pong 2050 & \SI{10.7}{\metre\per\second} & \SI{106.7}{\per\second} \\
AIMY (MN4004 KV300) & \SI{10.8}{\metre\per\second} & $-$\hspace{0.4em} \si{\per\second} \\
AIMY (MN5008 KV170) & \SI{15.4}{\metre\per\second} & \SI{192.0}{\per\second} \\ \hline
\end{tabular}
\end{table}

\subsection{Evaluation of High-fidelty}

\begin{figure}[ht]
    \centering
    \subfloat[Distance accuracy with low altitude angle]{
        \includegraphics[width=0.95\textwidth]{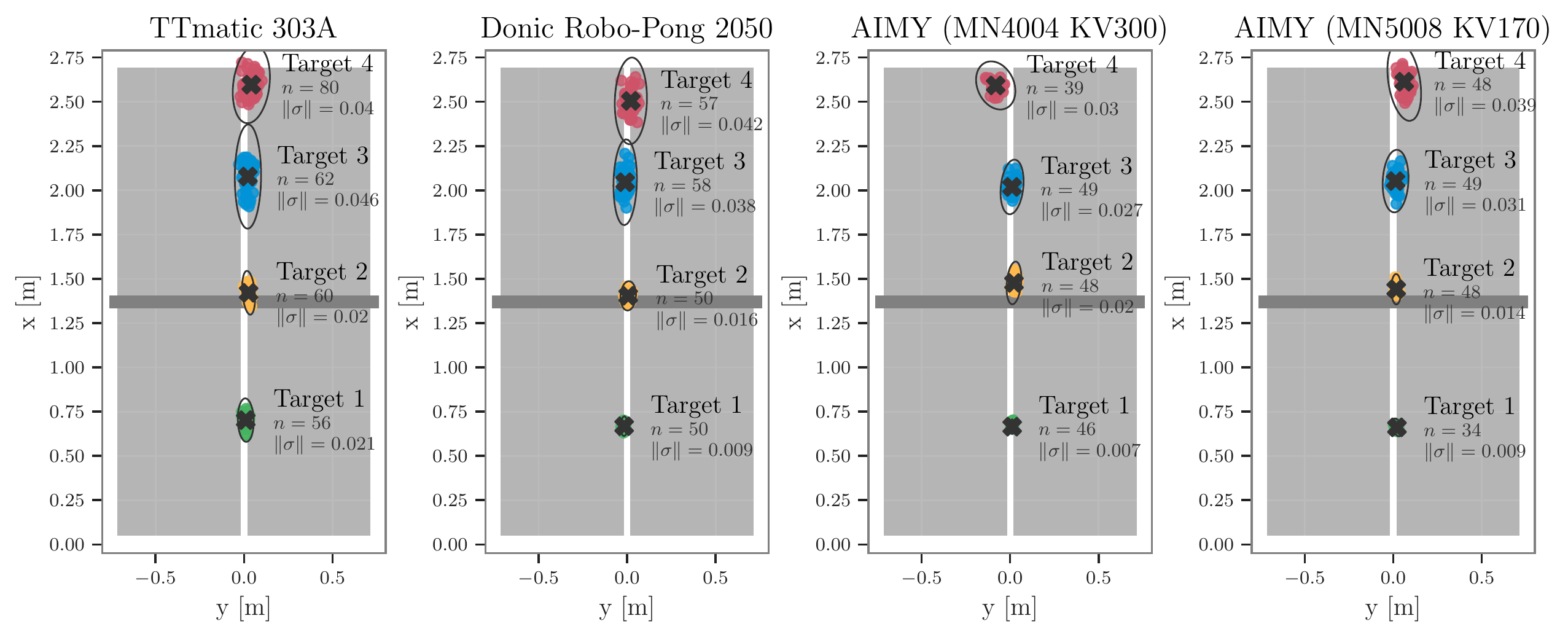}
        }
    \newline
    \subfloat[Distance accuracy with medium altitude angle]{
        \includegraphics[width=0.95\textwidth]{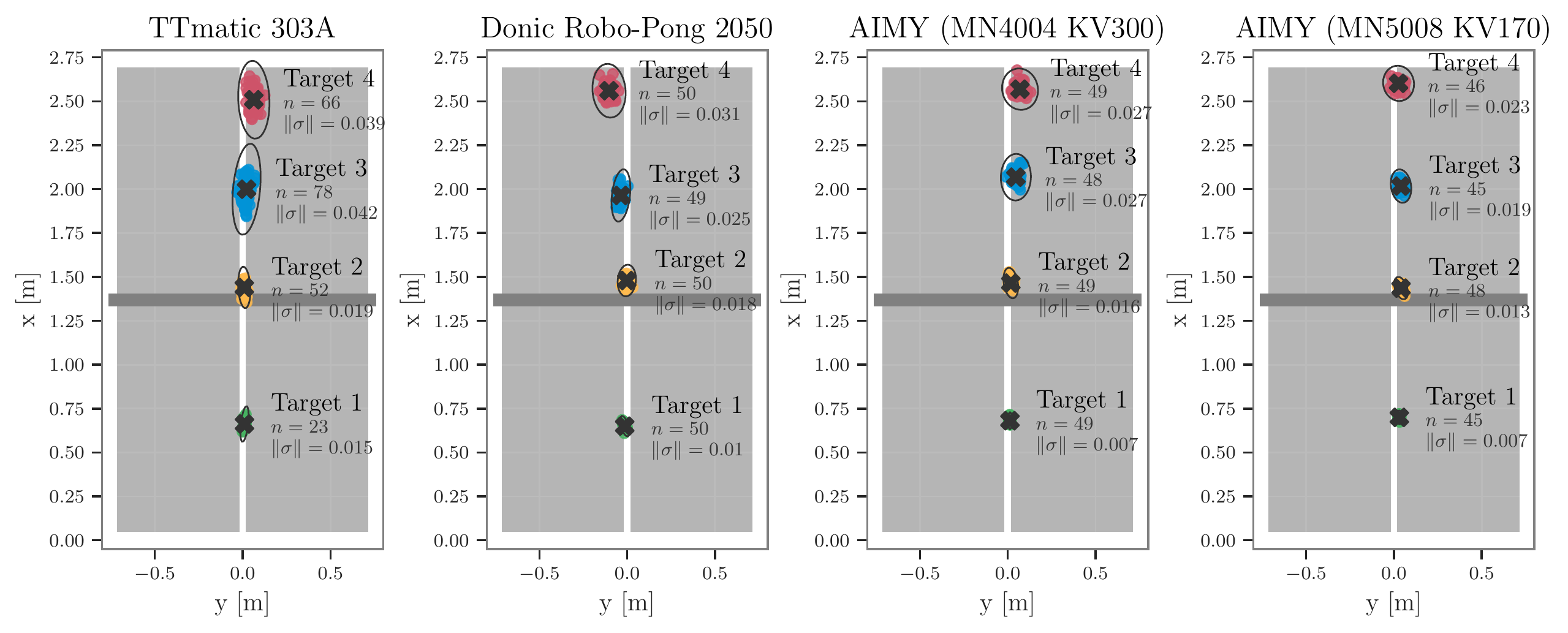}
        }
    \newline
    \subfloat[Distance accuracy with high altitude angle]{
        \includegraphics[width=0.95\textwidth]{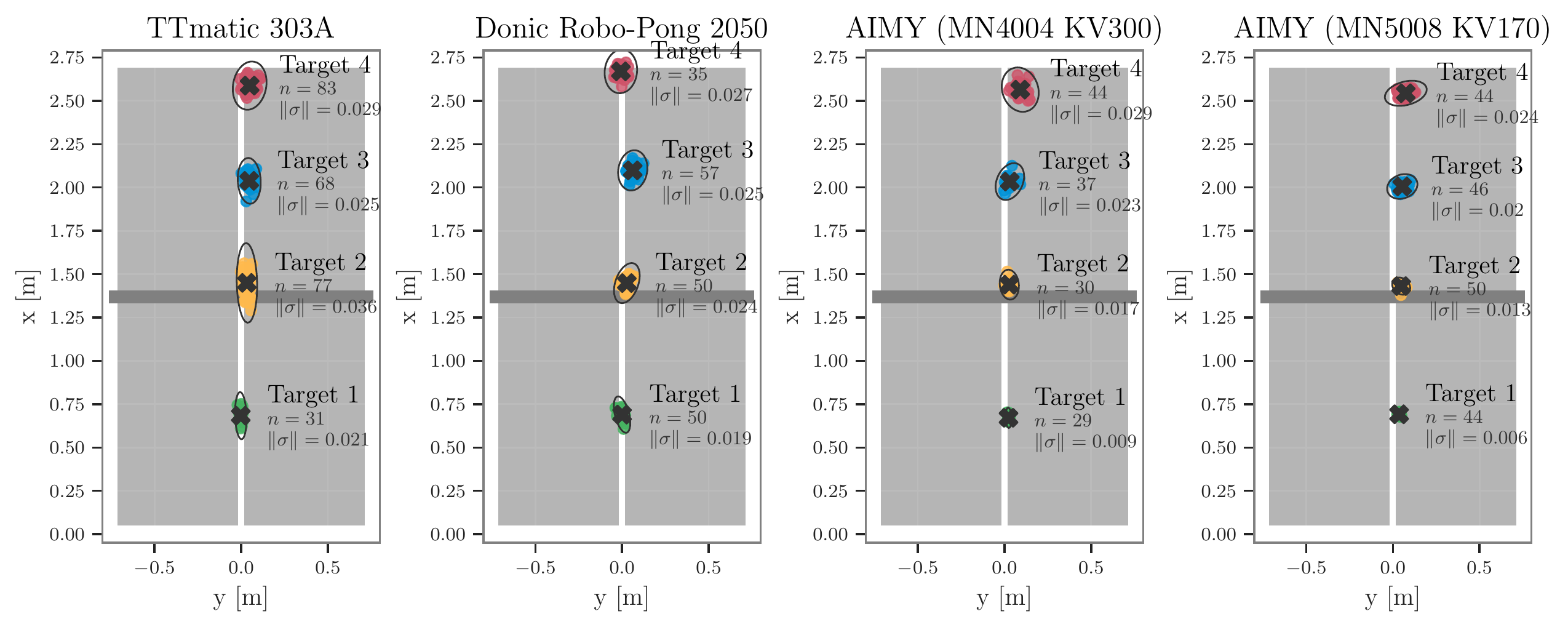}
        }
    \newline
    \caption{Distance accuracy experiment for low, medium, and high altitude angles.}
    \label{fig:appendix_launchers}
\end{figure}

In Figure~\ref{fig:launcher_comparison}, the accuracy of AIMY was compared to the TTmatic~303A and Donic Robo-Pong 2050 for four distances with the medium altitude setting. Table~\ref{tab:angles} lists the angle values for each altitude setting. Figure~\ref{fig:appendix_launchers} extends Figure~\ref{fig:launcher_comparison} with low and high altitude angles. 

\begin{table}[H]
\centering
\caption{Altitude angles of evaluated ball launchers}
\begin{tabular}{@{}rccc@{}}
\toprule
\multicolumn{1}{c}{} &
  \textbf{\begin{tabular}[c]{@{}c@{}}Low\\ Angle\end{tabular}} &
  \textbf{\begin{tabular}[c]{@{}c@{}}Medium\\ Angle\end{tabular}} &
  \textbf{\begin{tabular}[c]{@{}c@{}}High\\ Angle\end{tabular}} \\ \midrule
TTmatic 303A &
  \begin{tabular}[c]{@{}c@{}}Slide with smallest\\ curvature\end{tabular} &
  \begin{tabular}[c]{@{}c@{}}Slide with medium\\ curvature\end{tabular} &
  \begin{tabular}[c]{@{}c@{}}Slide with largest\\ curvature\end{tabular} \\
Donic Robo-Pong 2050 & \ang{8} & \ang{26} & \ang{30} \\
AIMY MN4004 KV300 & \ang{14} & \ang{26} & \ang{28} \\
AIMY MN5008 KV170 & \ang{6.4} & \ang{19.9} & \ang{37.1} \\ \bottomrule
\end{tabular}
\label{tab:angles}
\end{table}

\subsection{Data Set}

\begin{figure}[H]
    \centering
    \includegraphics[width=1.0\textwidth]{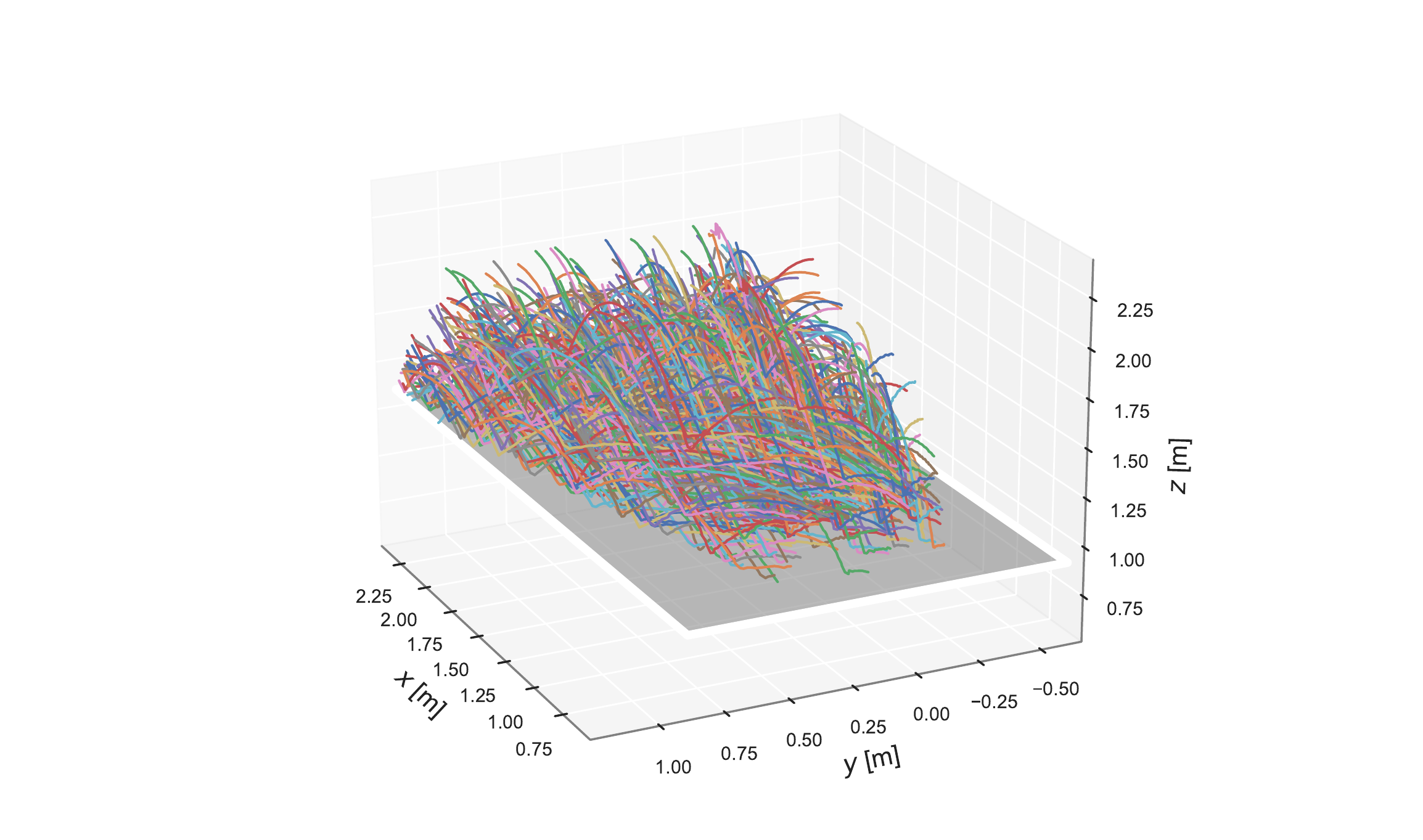}
    \caption{Ball trajectories from recorded training data set in three-dimensions. Grey plane illustrates the table tennis table. Ball trajectories start from the frontend of the table and cover the other side of the table, cone-shaped.}
    \label{fig:appendix_dataset3d}
\end{figure}

\begin{figure}[H]
    \centering
    \includegraphics[width=0.65\textwidth]{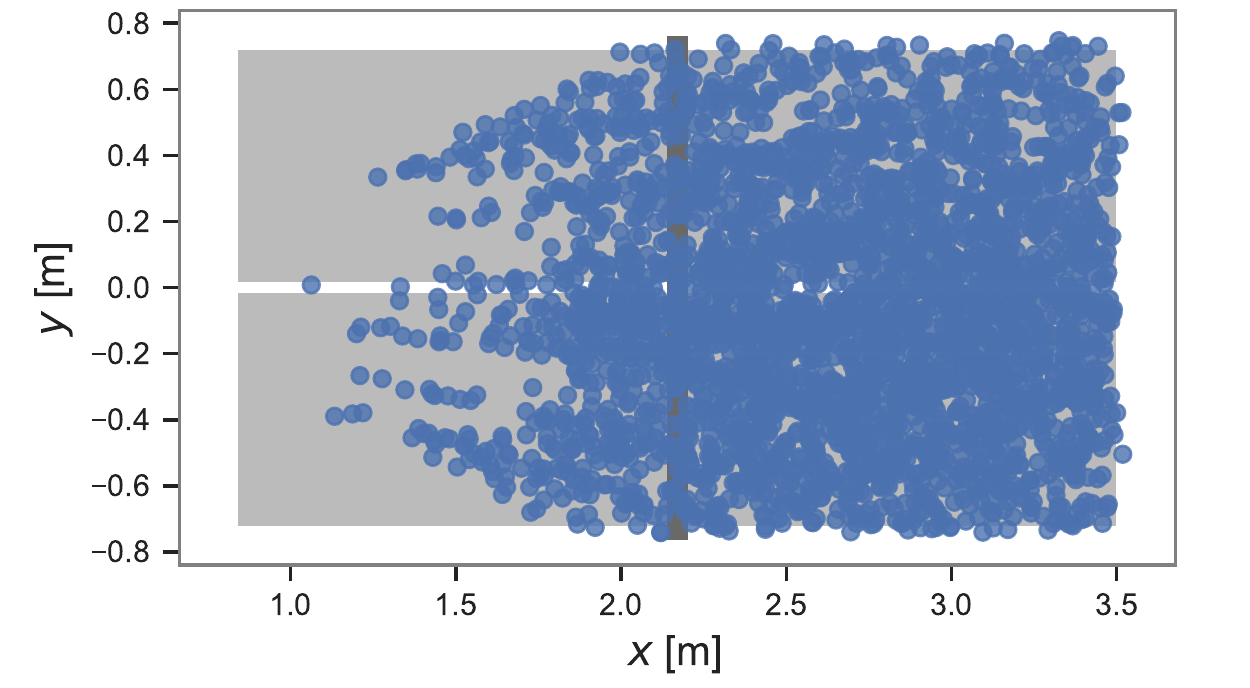}
    \caption{Landing points of the ball trajectories of the training data set. Grey plane illustrates the table tennis table. The opposite side of the table has a high coverage of landing points.}
    \label{fig:appendix_dataset_hitpoints}
\end{figure}

\begin{figure}[H]
    \centering
    \includegraphics[width=0.45\textwidth]{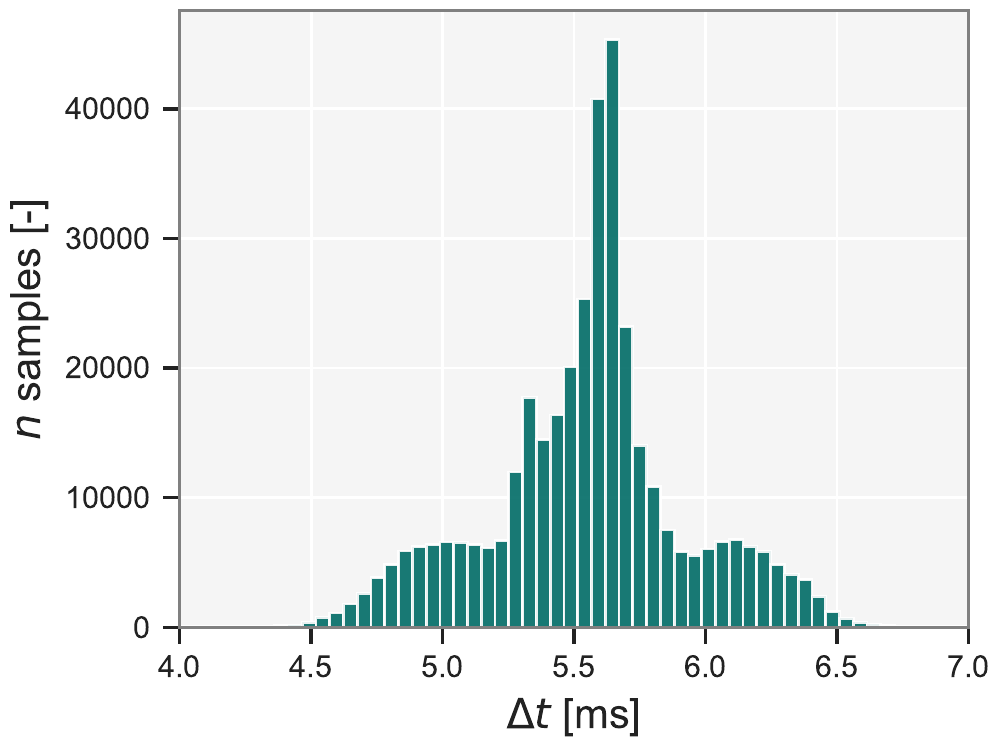}
    \caption{Histogram of the time delta between each recorded sample. The time differences are located between 4.5 and \SI{6.5}{\milli\second}. The majority of the time differences are between approximately 5.25 and \SI{5.75}{\milli\second}.}
    \label{fig:appendix_histogram}
\end{figure}

For illustration purposes, Figure~\ref{fig:appendix_dataset3d} shows 415 of the 3761 recorded trajectories for training after preprocessing. The landing points of the ball trajectories of the training data set are shown in Figure~\ref{fig:appendix_dataset_hitpoints}. The ball tracking system described in Section~\ref{sec:accuracy} records with \SI{200}{\hertz}. Figure~\ref{fig:appendix_histogram} depicts a histogram of the time stamp differences of consecutive samples to verify the equidistance of the recorded samples. We assume, that computation in the ball detection algorithm leads to slightly reduced recording frequency.

\subsection{Target Shooting}

For learning to hit landing position, we trained a feed-forward neural network, as described in Section~\ref{sec:target_eval}. For training, we consider each position measurement of the presented data set and the corresponding control parameters as a training sample. All training samples after the rebound are not considered. For target shooting, we simplified the problem by only considering trajectories shot with equal throwing wheel speeds. This simplification reduced the control parameters to the two orientation angles and one throwing wheel speed applied to all wheels, and herewith the dimensionality of the action space. Therefore, the target shooting network learns the position to control parameter mapping with a total of 66581 training samples from 415 recorded trajectories. Table~\ref{tab:nn_specs} lists the specifications of the architecture of the neural network.

\begin{table}[H]
\centering
\caption{Specification of the target shooting network}
\label{tab:nn_specs}
\begin{tabular}{@{}cc@{}}
\toprule
\multicolumn{1}{c}{\textbf{Specification}} & \multicolumn{1}{c}{\textbf{Value}} \\ \midrule
Number of Layers       & 3                                  \\
Neurons per Layer      & 2048, 512, 128                     \\
Loss                   & MSE                                \\
Activation             & sigmoid                            \\
Dropout                & 0.1                                \\
Optimizer              & ADAM                               \\
Epochs                 & 1400                               \\
Mini batch size        & 4096                               \\
Minimum learning rate  & 1e-4                               \\ 
Normalization          & standard                           \\ 
\bottomrule
\end{tabular}
\end{table}


\end{document}